\documentclass[10pt,twocolumn,letterpaper]{article}

\usepackage{cvpr}              

\usepackage{graphicx}
\usepackage{amsmath}
\usepackage{amssymb}
\usepackage{booktabs}

\usepackage[pagebackref,breaklinks,colorlinks]{hyperref}

\usepackage[capitalize]{cleveref}
\crefname{section}{Sec.}{Secs.}
\Crefname{section}{Section}{Sections}
\Crefname{table}{Table}{Tables}
\crefname{table}{Tab.}{Tabs.}

\usepackage{multirow}
\usepackage{rotating}
\usepackage{adjustbox}
\usepackage[dvipsnames]{xcolor}
\newcommand*{\red}{\textcolor{red}}
\newcommand*{\green}{\textcolor{green}}
\def\eg{\emph{e.g}\onedot} 
\def\ie{\emph{i.e}\onedot} 
\usepackage{mathtools}
\usepackage{textcomp}
\usepackage{gensymb}
\usepackage{enumitem}

\usepackage{amssymb}
\usepackage{pifont}
\newcommand{\cmark}{\ding{51}}
\newcommand{\xmark}{\ding{55}}
\newcommand\RX{\text{\red{\xmark}}}
\newcommand\GT{\text{\green{\cmark}}}

\newcommand\our{\text{CosPlace}}
\newcommand\ourD{\text{SF-XL}}
\newcommand\supplementary{\text{appendix}}

\newcommand{\myparagraph}[1]{\vspace{3pt}\noindent\textbf{#1}}

\usepackage{amsthm}

\def\cvprPaperID{6518} 
\def\confName{CVPR}
\def\confYear{2022}

\begin{document}

\title{Rethinking Visual Geo-localization for Large-Scale Applications}

\author{%
  \textbf{Gabriele Berton} \\
  Politecnico di Torino\\
  {\tt\small gabriele.berton@polito.it}
  \and
  \textbf{Carlo Masone}\\
  CINI
  \and
  \textbf{Barbara Caputo}\\
  Politecnico di Torino\\
}

\maketitle

\begin{abstract}
Visual Geo-localization (VG) is the task of estimating the position where a given photo was taken by comparing it with a large database of images of known locations.
To investigate how existing techniques would perform on a real-world city-wide VG application, we build San Francisco eXtra Large, a new dataset covering a whole city and providing a wide range of challenging cases, with a size 30x bigger than the previous largest dataset for visual geo-localization.
We find that current methods fail to scale to such large datasets, therefore we design a new highly scalable training technique, called {\our}, which casts the training as a classification problem avoiding the expensive mining needed by the commonly used contrastive learning.
We achieve state-of-the-art performance on a wide range of datasets and find that {\our} is robust to heavy domain changes.
Moreover, we show that, compared to the previous state-of-the-art, {\our} requires roughly 80\% less GPU memory at train time, and it achieves better results with 8x smaller descriptors, paving the way for city-wide real-world visual geo-localization.
Dataset, code and trained models are available for research purposes at 
{\small{\url{https://github.com/gmberton/CosPlace}}}.
\end{abstract}

\begin{figure}
    \centering
    \includegraphics[width=0.75\linewidth]{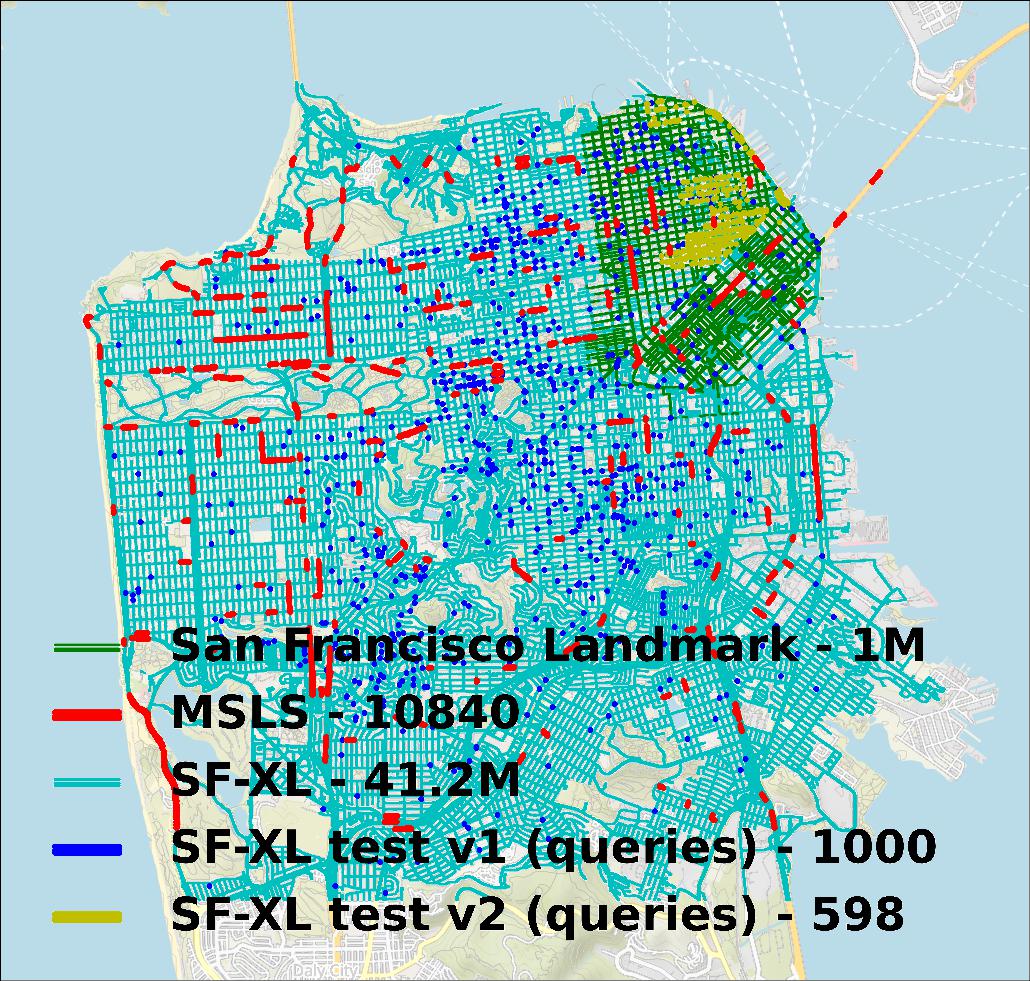}
    \caption{\textbf{Map of various datasets on the city of San Francisco.} Previous datasets only cover a sector of the city (green points) or are sparse (red points). {\ourD} densely covers the whole city and provides a realistic case-study for large-scale applications.}
    \label{fig:map}
    \vspace{-0.1cm}
\end{figure}

\section{Introduction}
\label{sec:introduction}
Visual geo-localization (VG), also known as visual place recognition \cite{Arandjelovic_2018_netvlad} or image localization \cite{Liu_2019_sare}, is a staple of computer vision \cite{Torii-2015,Arandjelovic_2018_netvlad,Torii_2018_tokyo247,Doan-2019,Hausler_2021_patch_netvlad,Torii_2021_r_sf,Zaffar-2021} and robotics research \cite{Chen-2017,Chen-2017b,Chen-2018b,Garg-2019,Khaliq-2020,Hausler-2019} and it is defined as the task of coarsely recognizing the geographical location where a photo was taken, usually with a tolerance of few meters \cite{Arandjelovic_2018_netvlad, Kim_2017_crn, Liu_2019_sare, Ge_2020_sfrs, Berton_2021_svox, Ibrahimi_2021_insideout_vpr, Warburg_2020_msls}.
This task is commonly approached as an image retrieval problem where the query to be localized is compared to a database of geo-tagged images: the most similar images retrieved from the database, together with their metadata, represent the hypotheses of the query's geographical location.
In particular, all recent VG methods are learning-based and use a neural network to project the images into an embedding space that well represents the similarity of their locations, and that can be used for the retrieval.

So far, research on VG has focused on recognizing the location of images in moderately sized geographical areas (\eg, a neighborhood). However, real-world applications of this technology, such as autonomous driving \cite{Doan-2019} and assistive devices \cite{Cheng-2020}, are posed to operate at a much larger scale (\eg, cities or metropolitan areas), thus requiring massive databases of geo-tagged images to execute the retrieval.
Having access to such massive databases, it would be advisable to use them also to train the model rather than just for the execution of the retrieval (inference).
This idea requires us to rethink VG, addressing the two following limitations.

\myparagraph{Non-representative datasets.}
The current datasets for VG are not representative of realistic large-scale applications, because they are either too small in the geographical coverage \cite{Arandjelovic_2018_netvlad, Torii_2018_tokyo247, Milford_2008_st_lucia, Geiger_2013_kitti, CarlevarisBianco_2016_nclt} or too sparse \cite{Warburg_2020_msls, Cummins_2009_eynsham, Maddern_2017_robotCar, Bansal_2014_cmu} (see \cref{fig:map} for an example of these limitations).
Moreover, current datasets follow the common practice of splitting the collected images into geographically disjoint sets for training and inference. However, this practice does not find a correspondence in the real world where one would likely opt to use images from the target geographical area to train the model. Considering also the cost of collecting the images, it would be advisable to use the whole database also for training.

\myparagraph{Scalability of training.}
Having access to a massive amount of data raises the question of how to use it effectively for training. All the recent state-of-the-art methods in VG use contrastive learning~\cite{Arandjelovic_2018_netvlad, Kim_2017_crn, Ge_2020_sfrs, Liu_2019_sare, Warburg_2020_msls, Hausler_2021_patch_netvlad, Peng_2021_appsvr, Peng_2021_sralNet, Berton_2021_svox, Ibrahimi_2021_insideout_vpr} (mostly relying on a triplet loss), which heavily depends on mining of negative examples across the training database \cite{Arandjelovic_2018_netvlad}. This operation is expensive, and it becomes prohibitive when the database is very large. Lightweight mining strategies that explore only a small pool of samples can reduce the duration of the mining phase \cite{Warburg_2020_msls}, but they still result in a slow convergence and possibly less effective use of the data.

\myparagraph{Contributions.}
In this paper, we address these two limitations with the following contributions:
\begin{itemize}
    \item A new large-scale and dense dataset, called San Francisco eXtra Large ({\ourD}), that is roughly 30x bigger than what is currently available (see \cref{fig:map}). The dataset includes crowd-sourced (\ie, multi-domain) queries that make for a challenging problem.
    \item A procedure that uses a classification task as a proxy to train the model that is used at inference to extract discriminative descriptors for the retrieval. We call this method {\our}. {\our} is remarkably simple, it does not necessitate to mine negative examples, and it can effectively learn from massive collections of data. 
\end{itemize}

Through extensive experimental validation, we demonstrate that not only {\our} requires roughly 80\% less GPU memory at train time than current SOTA, but also that a simple model trained with {\our} on {\ourD} surpasses the SOTA while using 8x smaller embeddings. Additionally, we show that this model generalizes far better to other datasets.

\section{Related work}
\label{sec:related_work}

\myparagraph{Visual geo-localization as image retrieval.}
Visual geo-localization is commonly approached as an image retrieval problem, with a retrieved image deemed correct if it is within a predefined range (usually 25 meters) from the query's ground truth position \cite{Arandjelovic_2018_netvlad, Kim_2017_crn, Liu_2019_sare, Ge_2020_sfrs, Berton_2021_svox, Warburg_2020_msls}.
All recent VG methods perform the retrieval using learned embeddings that are produced by a feature extraction backbone equipped with a head that implements some form of aggregation or pooling, the most notable being NetVLAD \cite{Arandjelovic_2018_netvlad}.
These architectures are trained via contrastive learning, typically using a triplet loss \cite{Arandjelovic_2018_netvlad, Kim_2017_crn, Warburg_2020_msls, Hausler_2021_patch_netvlad, Peng_2021_appsvr, Peng_2021_sralNet, Berton_2021_svox, Ibrahimi_2021_insideout_vpr} and leveraging the geo-tags of the database images as a form of weak supervision to mine negative examples \cite{Arandjelovic_2018_netvlad}. There are various alternatives to this scheme, such as the Stochastic Attraction-Repulsion Embedding (SARE) loss \cite{Liu_2019_sare}, which allows to efficiently use multiple negatives at once, while keeping the training methodology from \cite{Arandjelovic_2018_netvlad}.
Another solution is presented in \cite{Ge_2020_sfrs}, and it achieves new state-of-the-art results by computing the loss not only over the full images but also over cleverly mined crops.
Despite the variations, all these methods suffer from poor train-time scalability caused by expensive mining techniques.  
The problem of scalability also arises at test time, and it relates to the size of the descriptors, which impacts the required memory and the retrieval time.
Although the dimensionality of NetVLAD descriptors can be reduced using PCA, this leads to a degradation of results~ \cite{Arandjelovic_2018_netvlad}, hence works like \cite{Ge_2020_sfrs, Liu_2019_sare} prefer to keep a high dimensionality of 4096.
Alternatively, other works drop NetVLAD and instead use pooling \cite{Radenovic_2019_gem} to build smaller embeddings. 


\myparagraph{Visual geo-localization as classification.}
An alternative approach to visual geo-localization is to consider it a classification problem \cite{Weyand_2016_PlaNet, Seo_2018_CPlaNet, Muller_2018_hierarchical_geolocation, Izbicki_2020_earth_geoloc, Kordopatis_2021_rrm}. 
These works build on the idea that two images coming from the same geographical region, although representing different scenes, are likely to share similar semantics, such as architectural styles, types of vehicles, vegetation, \etc.
In practice, these methods divide the geographical area of interest in cells and group the database of images in classes according to their cell. This formulation allows to scale the problem to the whole globe, but at the cost of reduced accuracy in the estimates because each class can span many kilometers. Therefore, these methods are not used to perform geo-localization when the estimates require tolerance of a few meters.


\myparagraph{Relation to prior works.}
In this work, we propose a new approach to VG that combines the advantages of both retrieval (tolerance of a few meters) and classification (high scalability). 
Our method uses a classification task as a proxy to train the model without requiring any mining. This makes it possible to train with massive datasets, unlike the solutions that use contrastive learning~\cite{Arandjelovic_2018_netvlad, Kim_2017_crn, Warburg_2020_msls, Hausler_2021_patch_netvlad, Peng_2021_appsvr, Peng_2021_sralNet, Berton_2021_svox, Ibrahimi_2021_insideout_vpr}. At test time, the trained model is used to extract image descriptors and perform a classic retrieval.
While {\our} might appear similar to previous classification-based works~\cite{Weyand_2016_PlaNet, Seo_2018_CPlaNet, Muller_2018_hierarchical_geolocation, Izbicki_2020_earth_geoloc, Kordopatis_2021_rrm}, given that they also partition a map into classes, there are substantial differences.
These prior works tackle the task of global classification and group images within very large cells (up to hundreds of kilometers wide), building on the idea that nearer scenes have similar semantics (\eg if two images are from China, they might both depict Chinese ideograms). On the other hand, our partitioning strategy is designed to leverage the availability of dense data and ensure that if two images are from the same class, they visualize the same scene.
Moreover, unlike~\cite{Weyand_2016_PlaNet, Seo_2018_CPlaNet, Muller_2018_hierarchical_geolocation, Izbicki_2020_earth_geoloc, Kordopatis_2021_rrm}, once trained our method can be used to perform geo-localization through image retrieval on any given geographical area.

\section{The San Francisco XL dataset}
\label{sec:dataset}
There are numerous datasets for VG (see \cref{tab:datasets_comparison}), but none of them reflects the scenario in which the geo-localization must be performed in  a large environment and with few meters of tolerance:
some datasets are limited to a small geographical area 
\cite{Arandjelovic_2018_netvlad,Torii_2018_tokyo247,Berton_2021_svox,CarlevarisBianco_2016_nclt} 
whereas
others do not densely cover the area 
\cite{Warburg_2020_msls,Geiger_2013_kitti,Cummins_2009_eynsham,Maddern_2017_robotCar,Bansal_2014_cmu,Sattler_2012_aachen}.
The San Francisco Landmark Dataset \cite{Chen_2011_san_francisco} partially overcomes these limitations, but it does not cover the whole city, nor has long-term temporal variations, which are essential for robustly training a neural network \cite{Arandjelovic_2018_netvlad}. Here we propose the first city-wide, dense, and temporally variable dataset: San Francisco eXtra Large ({\ourD}).

\begin{table}[!t]
  \centering
  \begin{adjustbox}{width=\linewidth}
  \begin{tabular}{|c|c|c|c|c|c|c|c|c|c|}
  \hline
  \multirow{2}{*}{Dataset}
  & \multirow{2}{*}{\# images}
  & \multicolumn{2}{c|}{Scenario}
  & \multicolumn{4}{c|}{Appearance Changes}
  & \multirow{2}{*}{\begin{tabular}[c]{@{}l@{}}Dense\\Coverage\end{tabular}}
  & \multirow{2}{*}{\begin{tabular}[c]{@{}l@{}}6 DoF\\Poses\end{tabular}}
  \\
  \cline{3-8}
  &  & \begin{turn}{90}Urban\end{turn} & \begin{turn}{90}Suburban\end{turn}
  & \begin{turn}{90}Weather\end{turn} & \begin{turn}{90}Long-term\end{turn}
  & \begin{turn}{90}Day/Night\end{turn} & \begin{turn}{90}Intrinsics\end{turn}
  & & 
  \\
  \hline
  KITTI \cite{Geiger_2013_kitti}                        & 13k   &\GT&\GT&\RX&\RX&\RX&\RX&\RX&\GT\\
  Eynsham \cite{Cummins_2009_eynsham}                   & 48k   &\GT&\GT&\RX&\RX&\RX&\RX&\RX&\RX\\
  St Lucia \cite{Milford_2008_st_lucia}                 & 33k   &\RX&\GT&\RX&\RX&\RX&\RX&\GT&\RX\\
  NCLT \cite{CarlevarisBianco_2016_nclt}                & 3.8M  &\GT&\RX&\RX&\RX&\RX&\RX&\GT&\GT\\
  Oxford RobotCar \cite{Maddern_2017_robotCar}          & 27k   &\GT&\GT&\GT&\GT&\GT&\RX&\RX&\GT\\
  CMU \cite{Bansal_2014_cmu}                            & 128k  &\GT&\GT&\GT&\GT&\RX&\RX&\RX&\GT\\
  Pittsburgh250k \cite{Arandjelovic_2018_netvlad}       & 278k  &\GT&\RX&\RX&\GT&\RX&\RX&\GT&\RX\\
  TokyoTM/247 \cite{Torii_2018_tokyo247}                & 189k  &\GT&\RX&\RX&\GT&\GT&\GT&\GT&\RX\\
  MSLS \cite{Warburg_2020_msls}                         & 1.7M  &\GT&\GT&\GT&\GT&\GT&\GT&\RX&\RX\\
  San Francisco Landmark \cite{Chen_2011_san_francisco} & 1.1M  &\GT&\RX&\RX&\RX&\RX&\GT&\GT&\GT\\
  Aachen \cite{Sattler_2012_aachen}                     & 4k    &\GT&\RX&\RX&\RX&\GT&\GT&\RX&\GT\\
  \hline
  {\ourD} (Ours)         & 41.2M &\GT&\GT&\GT&\GT&\GT&\GT&\GT&\GT\\
  \hline
  \end{tabular}
  \end{adjustbox}
  \caption{\textbf{Comparison of various VG datasets.} 
  }
  \label{tab:datasets_comparison}
  \vspace{-0.3cm}
\end{table}

\myparagraph{Database.}
Like other datasets used in this task \cite{Arandjelovic_2018_netvlad, Torii_2018_tokyo247, Berton_2021_svox, Knopp_2010_geotagged_streetview, Zamir_2014_102k_streetview, Zemene_2017_world_cities}, {\ourD}'s database is created from Google StreetView imagery. We collected 3.43M equirectangular panoramas (360° images) and split them horizontally in 12 crops, following \cite{Torii_2018_tokyo247, Arandjelovic_2018_netvlad}. This results in a total of 41.2M images (some examples are presented in \cref{fig:examples}).
Each crop is labeled with 6~DoF information (which includes GPS and heading).
The images were taken between 2009 and 2021, thus providing an abundance of long-term temporal variations.

Besides scale and density, {\ourD} differs from previous datasets \cite{Arandjelovic_2018_netvlad,Torii_2018_tokyo247,Warburg_2020_msls,Berton_2021_svox} in that the database is not split in geographically non-overlapping subsets for training, validation and testing. 
We argue that such division of the database does not reflect the reality of applications that may use VG.
In fact, when tasked with building a VG application for a large geographical area, one would likely opt to train a neural network on the imagery of such area rather than additionally collecting images from a disjointed one (perhaps adjacent).

To this end, we use the 41.2M images as a training set, which therefore covers the whole area of San Francisco.
While at test time, we could use the whole database of 41.2M images for the retrieval, this would be prohibitive for research purposes because extracting the descriptors for all the images on a single GPU requires days.
Therefore, we use as test time database only the 2.8M images from the year 2013, which still cover the whole SF-XL geographical area, 
We found this choice to be a good solution because the set of images is small enough to be a feasible research option for testing but large enough to simulate a real-world scenario and prevent the bad practice of validating or tuning hyperparameters on the test set (as it would take too long to validate on it every epoch).

Finally, for validation, we use a small set of images scattered through the whole city, made of 8k database images and 8k queries.

\myparagraph{Queries.}
While previous methods require the train set to be split into database and queries, {\our} does not need this distinction.
For this reason, we release the training set as a whole, and we believe that methods relying on database/queries splits for training (\eg, \cite{Arandjelovic_2018_netvlad, Kim_2017_crn, Liu_2019_sare, Ge_2020_sfrs, Peng_2021_appsvr, Peng_2021_sralNet, Yu_2020_SPEVlad}) should choose the split that best suits their needs, given that this choice can heavily influence the results (see in \cref{tab:comp_all_ds_full_dim} the difference in results between the third and fourth row as an example of this phenomenon).


\begin{table}
    \centering
    \begin{adjustbox}{width=0.9\linewidth}
    \begin{tabular}{|l|cccc|}
    \hline
    & SF-XL   & SF-XL & SF-XL test v1 & SF-XL test v2 \\
    & (Train) & (Val) & (Test)        & (Test)   \\
    \hline
    Database & \multirow{2}{*}{41.2M} & 8k & 2.8M  & 2.8M \\
    Queries  &                        & 8k & 1000  & 598 \\
    \hline
    \end{tabular}
    \end{adjustbox}
    \caption{\textbf{Number of images for each subset of {\ourD}}. The train set is not split into database and queries.}
    \label{tab:sf_xl_size}
\end{table}

\begin{figure}
    \centering
    \begin{minipage}{.12\textwidth}
        \includegraphics[width=\textwidth]{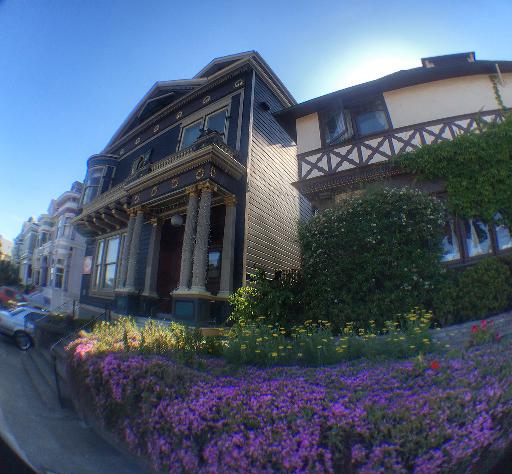}
    \end{minipage}
    \begin{minipage}{.17\textwidth}
        \includegraphics[width=\textwidth]{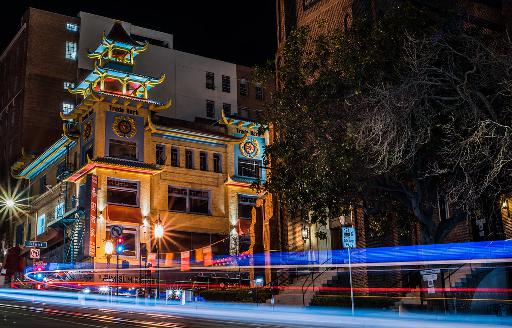}
    \end{minipage}
    \begin{minipage}{.15\textwidth}
        \includegraphics[width=\textwidth]{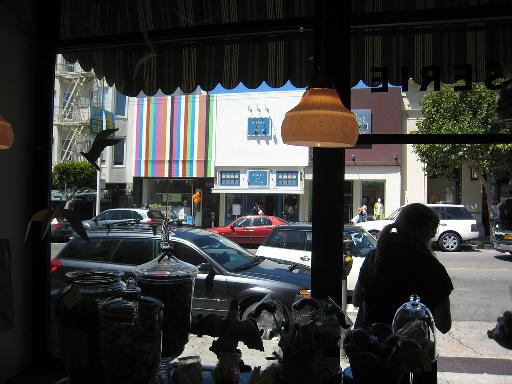}
    \end{minipage}
    \vspace{-0.2cm}
    \caption{\textbf{Examples of queries from SF-XL's test set v1}}
    \label{fig:queries}
    \vspace{-0.3cm}
\end{figure}

Concerning the test queries, we believe that they should not be from the same domain as the database, because in most real-world scenarios test time queries could come from unseen domains. This is also in agreement with what is done in the San Francisco Landmark Dataset \cite{Chen_2011_san_francisco} and Tokyo 24/7 \cite{Torii_2018_tokyo247}.
Hence, in {\ourD} we include two different sets of test queries:
\begin{itemize}[itemsep=3pt,topsep=3pt]
\item \textbf{test set v1}: a set of 1000 images collected from Flickr (similarly to Oxford5k \cite{Philbin_2007_oxford5k}, Paris6k \cite{Philbin_2008_paris6k}, Sfm120k \cite{Radenovic_2019_gem}). Given the inaccurate GPS coordinates of Flickr imagery, all the images in this set were hand-picked, and their location has been manually verified. We also made sure to blur out faces and plate numbers to anonymize the pictures. These images are very diverse and have a wide range of viewpoint and illumination (day/night) changes (see \cref{fig:queries});
\item \textbf{test set v2}: a set of 598 images from the queries of the San Francisco Landmark Dataset \cite{Chen_2011_san_francisco}, for which the 6~DoF coordinates have been generated by \cite{Torii_2021_r_sf}. Given that it provides 6~DoF labels, this set can also be used for large-scale pose estimation.
\end{itemize}

\cref{tab:sf_xl_size} presents a summary of {\ourD}, while further information about it are in \cref{sec:further_information_on_sf_xl}.

\section{Method}
\label{sec:method}
Given that a city-wide dataset requires millions of images (\cref{sec:dataset}) and that visual geo-localization is by nature a large-scale task, we believe that a proper method for VG should be highly scalable, both at train and at test time.
We find that current (and previous) state of the art \cite{Arandjelovic_2018_netvlad, Kim_2017_crn, Liu_2019_sare, Ge_2020_sfrs} lacks these important qualities:
\begin{itemize}[itemsep=3pt,topsep=3pt]
    \item at train time, they require to periodically compute the features of all database images and keep them in a cache: this results in a space and time complexity of $\mathcal{O}(n)$, which is suitable only for small datasets;
    \item these methods \cite{Arandjelovic_2018_netvlad, Kim_2017_crn, Liu_2019_sare, Ge_2020_sfrs} rely on a NetVLAD layer \cite{Arandjelovic_2018_netvlad} or its variants \cite{Kim_2017_crn}, which produce high dimensional vectors and require large amounts of memory for inference (given that database descriptors should be kept in memory for efficient retrieval).
    As an example, a VGG-16 with NetVLAD produces vectors of dimension 32k, which for a 10M database would require $32\text{k} \cdot 10\textrm{M} \cdot 4\textrm{B} = 1220\textrm{GB}$ of memory.
    Smaller embeddings are usually obtained with reduction techniques such as PCA, although using dimensions lower than 4096 leads to a rapid decline of results \cite{Arandjelovic_2018_netvlad, Peng_2021_appsvr, Peng_2021_sralNet, Zhu_2018_apanet}.
\end{itemize}

To reduce train time complexity we take inspiration from the domain of face recognition, where cosFace \cite{Wang_2018_cosFace} and arcFace \cite{Deng_2019_arcFace} are key in achieving state-of-the-art results \cite{Srivastava_2019_face_bench}.
These losses require the training set to be divided into classes; however, in VG the label space is continuous (GPS coordinates and optionally heading/orientation information), making it not straightforward to divide it into discrete classes.


\begin{figure}
    \centering
    \includegraphics[width=0.9\linewidth]{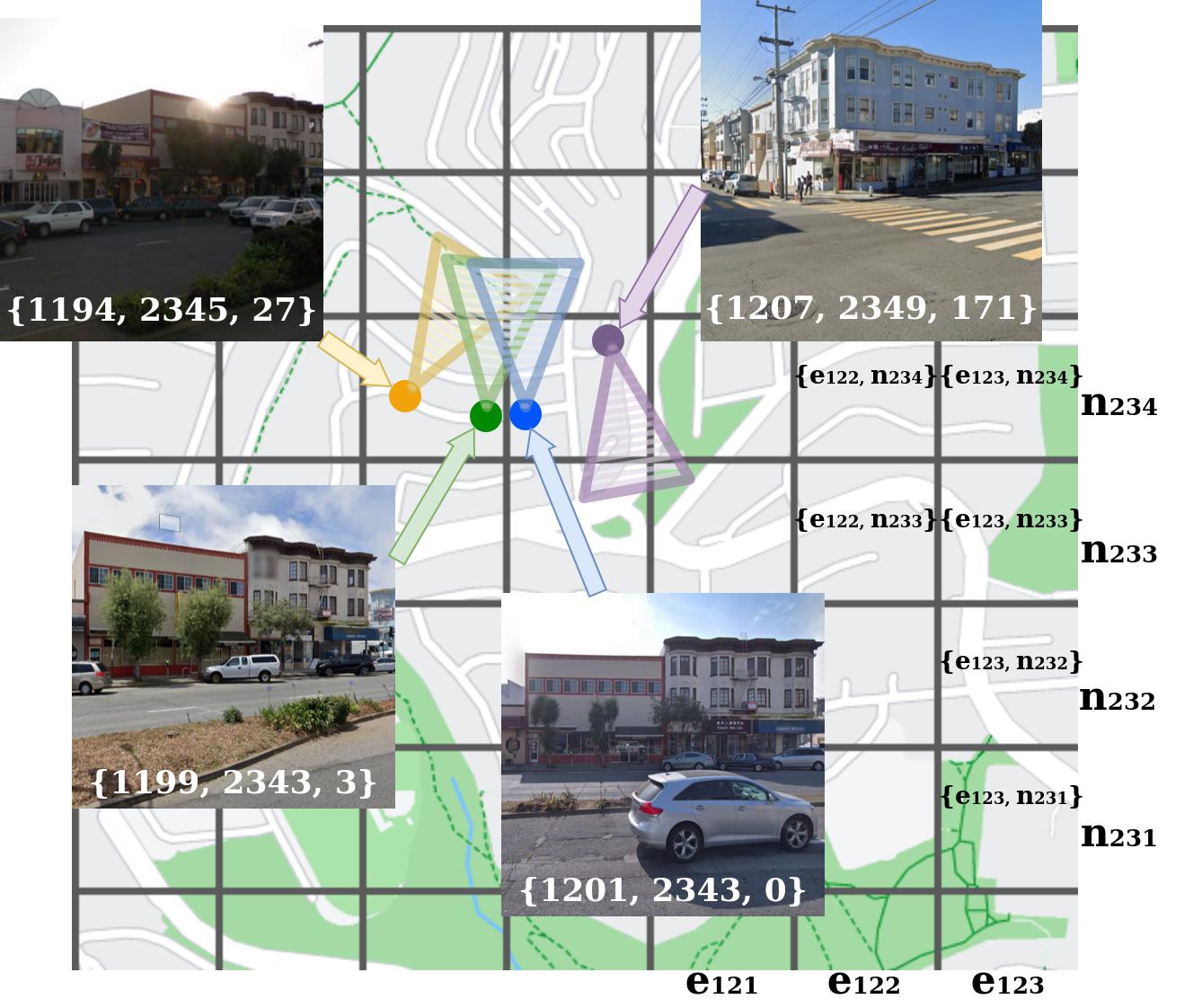}
    \vspace{-0.1cm}
    \caption{
    \textbf{Grid of the map showing how cells are formed based on UTM coordinates.}
    Each cell (of side lenght $M$) is identified by a pair of values (for readability only a few are shown on the right).
    On each of the four images is written its triplet $\{\textit{UTM east}, \textit{UTM north}, \textit{heading}\}$ in white between braces.
    Within any given cell there are images with very different headings/orientations (see the two images on the right), which should belong to different classes, hence the split of each cell into multiple classes according to their heading.
    We also see that the two images at the bottom, although representing the same scene/building, belong to different cells (and therefore to different classes), and this would be confusing for a naive classification-based algorithm.
    }
    \label{fig:map_grid}
    \vspace{-0.3cm}
\end{figure}

\subsection{Splitting the dataset into classes}
A naive approach to divide the database into classes would be to split it into square geographical cells (see \cref{fig:map_grid}), using UTM coordinates $\{\textit{east}, \textit{north}\}$\footnote{UTM coordinates are defined by a system used to identify locations on earth in meters, where 1 UTM unit corresponds to 1 meter. They can be extracted from GPS coordinates (\ie, latitude and longitude) and allow approximating a restricted area of the earth's surface on a flat surface.}, and further slice each cell into a set of classes according to each image's orientation/heading $\{\textit{heading}\}$.
Formally, the set of images assigned to the class $C_{e_i, n_j, h_k}$ would be 
\vspace{-3px}
\begin{equation}
    \label{eq:classes}
\scalebox{1}{$
    \left\{ x : \left\lfloor\frac{\textit{east}}{M} \right\rfloor = e_i, \left\lfloor\frac{\textit{north}}{M}\right\rfloor = n_j, \left\lfloor\frac{\textit{heading}}{\alpha}\right\rfloor = h_k\right\}$
}
\end{equation}
where $M$ and $\alpha$ are two parameters (respectively in meters and degrees) that determine the extent of each class in position and heading. 
While this solution creates a set of classes from a VG dataset, it has a big limitation:
nearly identical images (\eg, taken with the same orientation but a few centimeters apart) may be assigned to different classes due to quantization errors (see \cref{fig:map_grid}), which would confuse a classification-based training algorithm.

\begin{figure}
    \centering
    \includegraphics[width=1.0\columnwidth]{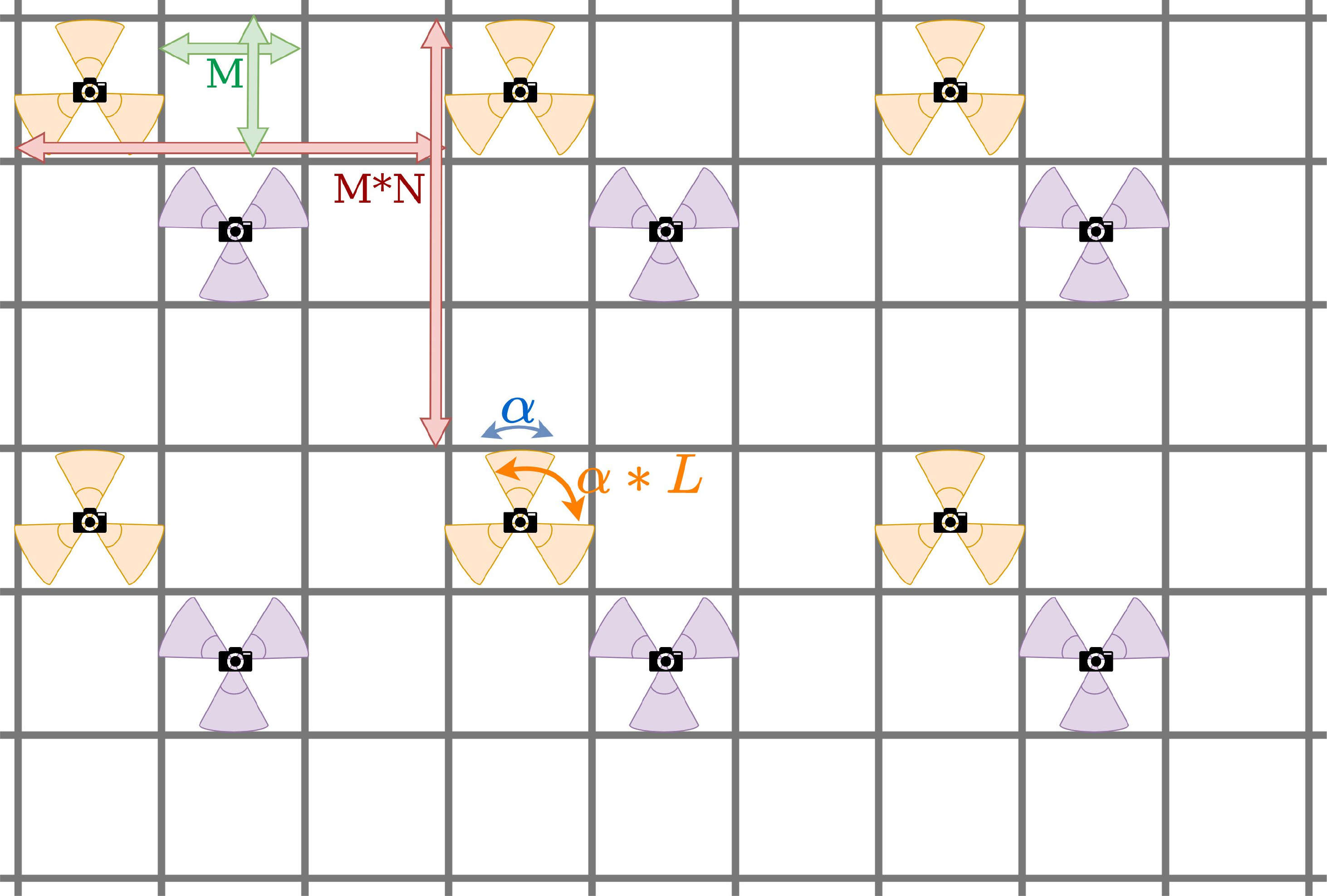}
    \caption{\textbf{Visual representation of {\our} groups.}
    The orange triangles represent one of the $N \times N \times L$ different groups, while the purple triangles represent another group.
    For clarity, only two groups are shown (namely $G_{000}$ and $G_{111}$).
    Each triangle represents one class, which contains all images within the respective cell with the proper orientation.
    The meaning for each of the four hyperparameters that define how groups are built (\ie, $M$, $\alpha$, $N$ and $L$) is visually shown.
    In the figure, $\alpha=60\degree, N=3, L=2$. The total number of groups with this configuration is therefore $3 \times 3 \times 2 = 18$, and each cell contains 6 classes from 2 groups.
    }
    \label{fig:map_groups}
    \vspace{-0.3cm}
\end{figure}

To overcome this limitation, we propose not to train the model using all the classes at once, but just groups of non-adjacent classes (two classes are adjacent if an infinitesimal difference in position or heading can bring an image from one class to the other).
Intuitively, these groups, which we call \textbf{{\our} Groups}, are akin to separate datasets and our proposed training procedure (explained later in \cref{sec:training}) iterates over them, one at a time.
We generate groups by fixing the minimum spatial separation that two classes of the same group should have, either in terms of translation or orientation.
For this purpose, we introduce two hyperparameters: $N$ controls the minimum number of cells between two classes of the same group, and $L$ is the equivalent for the orientation (see \cref{fig:map_groups} for a visual description).
Formally, we define the {\our} Group $G_{u v w}$ as the set of classes defined as follows
\vspace{-1px}
\begin{equation}
\begin{aligned}
\scalebox{0.95}{$G_{u v w}$} 
&\scalebox{0.95}{$= \big\{ C_{e_i n_j h_k} : (e_i \bmod N = u) \; \land $
}\\
& \scalebox{0.95}{$
     \land \; (n_j \bmod N = v) \land \left(h_k \bmod L = w\right) \big\}$    
}
\end{aligned}
\end{equation}
By construction, {\our} Groups are disjoint sets of classes from \cref{eq:classes} with the following properties:\\
\textbf{Property 1:} each class belongs to exactly one group.\\
\textbf{Property 2:} within a given group, if two images belong to different classes they are at least $M\cdot(N-1)$ meters apart \textbf{or} $\alpha \cdot (L-1)$ degrees apart (see \cref{fig:map_groups});\\
\textbf{Property 3:} the total number of {\our} Groups is $N \times N \times L$. \\
\textbf{Property 4:} no two adjacent classes can belong to the same group (unless $N=1$ or $L=1$).

\subsection{Training the network}
\label{sec:training}
We will now describe how the {\our} Groups are used to train the model.
We take inspiration from the Large Margin Cosine Loss (LCML) \cite{Wang_2018_cosFace} also known as cosFace \cite{Wang_2018_cosFace, Srivastava_2019_face_bench, Weyand_2020_gldv2}, which has shown remarkable results in face recognition \cite{Srivastava_2019_face_bench} and landmark retrieval \cite{Weyand_2020_gldv2}.
Nonetheless, vanilla LCML cannot be applied directly to any VG dataset, given that images are not split into a finite number of classes.
However, with the proposed partitioning of the dataset, we can perform LCML sequentially over each {\our} group (where each group can be considered as a separate dataset) and iterate over the many groups.
We call this training procedure \emph{CosPlace}.
As the LCML relies on a fully connected layer with output dimensionality equal to the number of classes, {\our} requires one fully connected layer per group: these layers are then discarded for validation/test.
Note that not all groups must necessarily be used for training, as one could simply use a single group and ignore the others; however, the full ablation at \cref{fig:ablation_full} shows that using more than one group helps achieve better results.
Therefore, we train sequentially over the groups. Formally, for each group:
\begin{equation}
    \mathcal{L}_\textit{cosPlace} = \mathcal{L}_\textit{lmcl}(G_{uvw})
\end{equation}
where $\mathcal{L}_\textit{lmcl}$ is the LCML loss as defined in \cite{Wang_2018_cosFace}, $u \in \{0, \ldots,N\}$, $v \in \{0, \ldots,N\}$, $W \in \{0,\ldots,L\}$.
In practice, we train one epoch on $G_{000}$, the next one on $G_{001}$, and so on, iterating over the groups.
The remarkable advantage of this procedure w.r.t. the methods based on contrastive losses commonly used in visual geo-localization, is that no mining nor caching is required, making it a much more scalable option.
At validation and test time, we use the model not to classify the query, but rather to extract image descriptors as in \cite{Wang_2018_cosFace} for a classic retrieval over the database.
This allows for the model to be used also on other datasets from unseen geographical areas (see \cref{tab:comp_all_ds_full_dim}).

\section{Experiments}
\label{sec:experiments}

\subsection{Implementation details}
\label{sec:implementation_details}

\myparagraph{Architecture.}
{\our} is architecture-agnostic, \ie it can be applied on virtually any image-based model.
For most experiments, we rely on a simple network made of a standard CNN backbone followed by a GeM pooling and a fully connected layer with output dimension 512.
Note that such a simple architecture is in contrast with the trend of the last five years of research in Visual Geo-localization \cite{Arandjelovic_2018_netvlad, Kim_2017_crn, Liu_2019_sare, Ge_2020_sfrs, Peng_2021_appsvr, Yu_2020_SPEVlad, Peng_2021_sralNet}, where the architectures rely on a more complex (w.r.t. our architecture) NetVLAD layer \cite{Arandjelovic_2018_netvlad}, and some even add a number of blocks on top of it \cite{Kim_2017_crn, Peng_2021_appsvr, Peng_2021_sralNet}, resulting in heavier and slower architectures.
Given that previous methods rely on a VGG-16 backbone \cite{Arandjelovic_2018_netvlad, Kim_2017_crn, Liu_2019_sare, Ge_2020_sfrs, Peng_2021_appsvr, Peng_2021_sralNet, Yu_2020_SPEVlad, Zhu_2018_apanet}, we use a VGG-16 in \cref{sec:comparison}, for fair comparisons with other methods.
For our ablations and preliminary experimentations, we relied on a ResNet-18, which achieves similar results to the VGG-16 at a fraction of the training time and memory requirements.
In \cref{sec:backbones} and \cref{sec:further_backbones}, we investigate the use of more recent backbones (ResNets \cite{He_2016_resnet} and transformer-based \cite{Dosovitskiy_2021_vit, Hassani_2021_cct}) with varying output dimensionality, finding that {\our} reaches encouraging results with a wide variety of architectures and thus demonstrates a great flexibility.
For example, we find that {\our} with a ResNet-101 backbone and 128-D descriptors outperforms current SOTA, which uses 4096-D descriptors.

\myparagraph{Training.}
Regarding hyperparameters, we set $M = 10$ meters, $\alpha = 30\degree$, $N = 5$ and $L = 2$.
We define an epoch as 10k iterations over a group.
We perform an epoch over the first group, then an epoch over the second group and so on, for a total of 50 epochs (\ie, 500k iterations with batch size of 32).
To ensure that the same group is seen more than once during training, we use only 8 of the $N \times N \times L$ (\ie, $5 \times 5 \times 2 = 50$) groups.
Validation is performed after each epoch, and once training is finished, testing is performed using the model that obtained the best performance on the validation set.
In \cref{sec:further_implementation_details}, we provide further implementation details, and we discuss how {\our} not only reduces the number of hyperparameters in comparison to previous VG methods but also that the ones it introduces have an intuitive meaning (see \cref{fig:map_groups}).

With respect to NetVLAD-based methods \cite{Arandjelovic_2018_netvlad, Kim_2017_crn, Liu_2019_sare, Ge_2020_sfrs, Peng_2021_appsvr, Yu_2020_SPEVlad, Peng_2021_sralNet}, which represent the entirety of the state of the art of the past five years, our method does not require multiple steps, whereas the NetVLAD layer requires for features clusters to be computed before starting to train the network, and, optionally, PCA to be computed afterwards.


\subsection{Comparison with other methods}
\label{sec:comparison}
In this section, we compare {\our} with previous methods in visual geo-localization.
For a proper assessment of the results, we test on 7 datasets, namely Pitts250k \cite{Arandjelovic_2018_netvlad}, Pitts30k \cite{Arandjelovic_2018_netvlad}, Tokyo 24/7 \cite{Torii_2018_tokyo247}, St Lucia \cite{Milford_2008_st_lucia}, Mapillary Street Level Sequences (MSLS) \cite{Warburg_2020_msls}, and our proposed SF-XL test v1 and SF-XL test v2.
Of these datasets, MSLS and St Lucia are composed of frontal view images taken with cars, while the rest rely on Google StreetView imagery to build the database.
For MSLS, given that the test set labels have not been publicly released yet, we test on the validation set as in \cite{Hausler_2021_patch_netvlad, Berton_2022_benchmark}.
As a metric, we use the recall@N with a 25 meters threshold, \ie, the percentage of queries for which at least one of the first N predictions is within a 25 meters distance from the query, following standard procedure \cite{Arandjelovic_2018_netvlad, Kim_2017_crn, Liu_2019_sare, Ge_2020_sfrs, Berton_2021_svox, Hausler_2021_patch_netvlad, Berton_2021_geowarp, Warburg_2020_msls}.

We perform experiments with NetVLAD \cite{Arandjelovic_2018_netvlad}, CRN \cite{Kim_2017_crn}, GeM \cite{Radenovic_2019_gem}, SARE \cite{Liu_2019_sare} and SFRS \cite{Ge_2020_sfrs}.
These methods are trained on the popular Pitts30k \cite{Arandjelovic_2018_netvlad} dataset, and we apply color jittering for all, following SFRS.
Although for some of them, it is intractable to train on a large-scale dataset such as MSLS or {\ourD} (\eg, SFRS's code\footnote{https://github.com/yxgeee/OpenIBL} relies on pairwise distance computation in features space between any pair of query-database image, which makes space and time complexity grow quadratically with the number of images), we are able to train GeM and NetVLAD also on MSLS and {\ourD} using the partial mining from \cite{Warburg_2020_msls}, which scales to larger datasets.
With this mining technique, we perform training with two different database/queries splits of {\ourD} (note that previous methods, unlike {\our}, require the dataset to be split):
{\ourD}* uses images after 2010 as queries, and the rest as database,
while {\ourD}** uses images after 2015 as queries and before 2015 as database (resulting in a denser database).
We also report results with other methods such as SPE-VLAD \cite{Yu_2020_SPEVlad}, SRALNet \cite{Peng_2021_sralNet}, APPSVR \cite{Peng_2021_appsvr} and APANet \cite{Zhu_2018_apanet}, although the code for these works is not publicly available and we could not independently reproduce the results.

\begin{table*}
\begin{adjustbox}{width=\textwidth}
\centering
\begin{tabular}{lccccccccccccccccccccccccc}
\toprule
\multicolumn{1}{l}{\multirow{2}{*}{Method}} & \multicolumn{1}{c}{\multirow{2}{*}{Desc. dim.}} & \multicolumn{1}{c}{\multirow{2}{*}{Train set}} & \multicolumn{2}{c}{Pitts250k} & & \multicolumn{2}{c}{Pitts30k} & & \multicolumn{2}{c}{Tokyo 24/7} & & \multicolumn{2}{c}{MSLS} & & \multicolumn{2}{c}{St Lucia} & & \multicolumn{2}{c}{Average} \\
\cline{4-5} \cline{7-8} \cline{10-11} \cline{13-14} \cline{16-17} \cline{19-20}
\multicolumn{3}{c}{}
& R@1  & R@5  & & R@1  & R@5  & & R@1  & R@5 &
& R@1  & R@5  & & R@1  & R@5  & & R@1  & R@5 \\
\hline
NetVLAD \cite{Arandjelovic_2018_netvlad} & 32768 & Pitts30k
& 85.9\footnotesize{$\pm$ 0.3} & 93.6\footnotesize{$\pm$ 0.2} &
& 86.1\footnotesize{$\pm$ 0.1} & 94.1\footnotesize{$\pm$ 0.3} &
& 62.2\footnotesize{$\pm$ 0.7} & 75.4\footnotesize{$\pm$ 0.5} &
& 54.8\footnotesize{$\pm$ 1.1} & 66.6\footnotesize{$\pm$ 1.0} &
& 70.8\footnotesize{$\pm$ 0.5} & 81.8\footnotesize{$\pm$ 0.7} 
& & 72.0 & 82.3 \\
NetVLAD \cite{Arandjelovic_2018_netvlad} & 32768 & MSLS
& 79.7\footnotesize{$\pm$ 0.8} & 90.2\footnotesize{$\pm$ 0.9} &
& 80.9\footnotesize{$\pm$ 1.4} & 90.6\footnotesize{$\pm$ 1.2} &
& 63.6\footnotesize{$\pm$ 1.1} & 77.5\footnotesize{$\pm$ 1.2} &
& 75.4\footnotesize{$\pm$ 0.4} & 84.2\footnotesize{$\pm$ 0.2} &
& 92.8\footnotesize{$\pm$ 0.4} & \textbf{97.6\footnotesize{$\pm$ 0.3}}
& & 78.5 & 88.0 \\
NetVLAD \cite{Arandjelovic_2018_netvlad} & 32768 & {\ourD}*
& 81.5\footnotesize{$\pm$ 0.1} & 90.8\footnotesize{$\pm$ 0.3} &
& 82.1\footnotesize{$\pm$ 0.2} & 91.5\footnotesize{$\pm$ 0.2} &
& 66.3\footnotesize{$\pm$ 0.4} & 78.2\footnotesize{$\pm$ 0.5} &
& 59.3\footnotesize{$\pm$ 0.8} & 68.7\footnotesize{$\pm$ 0.7} &
& 80.6\footnotesize{$\pm$ 0.5} & 90.7\footnotesize{$\pm$ 0.8}
& & 74.0 & 84.0 \\
NetVLAD \cite{Arandjelovic_2018_netvlad} & 32768 & {\ourD}**
& 76.2\footnotesize{$\pm$ 0.3} & 88.8\footnotesize{$\pm$ 0.4} &
& 78.8\footnotesize{$\pm$ 0.4} & 90.0\footnotesize{$\pm$ 0.4} &
& 56.2\footnotesize{$\pm$ 0.8} & 67.7\footnotesize{$\pm$ 0.9} &
& 51.4\footnotesize{$\pm$ 0.9} & 60.7\footnotesize{$\pm$ 0.7} &
& 78.5\footnotesize{$\pm$ 0.6} & 88.3\footnotesize{$\pm$ 0.8}
& & 68.2 & 79.1 \\
CRN \cite{Kim_2017_crn} & 32768 & Pitts30k
& 87.0\footnotesize{$\pm$ 0.4} & 94.5\footnotesize{$\pm$ 0.1} &
& 86.3\footnotesize{$\pm$ 0.4} & 94.6\footnotesize{$\pm$ 0.4} &
& 62.8\footnotesize{$\pm$ 0.5} & 77.4\footnotesize{$\pm$ 0.8} &
& 57.6\footnotesize{$\pm$ 0.3} & 70.4\footnotesize{$\pm$ 0.8} &
& 70.9\footnotesize{$\pm$ 0.7} & 82.8\footnotesize{$\pm$ 0.7}
& & 72.9 & 83.9 \\
SPE-VLAD \cite{Yu_2020_SPEVlad} \textdagger & 32768 & Pitts30k
&  -   &  -   & &  -   & 89.2 & &  -   & 63.9 &
&  -   &  -   & &  -   &  -   & &  -   &  -   \\
SARE \cite{Liu_2019_sare} &  4096 & Pitts30k
& 88.0\footnotesize{$\pm$ 0.6} & 94.8\footnotesize{$\pm$ 0.3} &
& 87.2\footnotesize{$\pm$ 0.8} & 93.9\footnotesize{$\pm$ 0.3} &
& 74.8\footnotesize{$\pm$ 3.0} & 84.3\footnotesize{$\pm$ 1.6} &
& 62.4\footnotesize{$\pm$ 2.9} & 73.2\footnotesize{$\pm$ 2.8} &
& 72.7\footnotesize{$\pm$ 2.6} & 86.0\footnotesize{$\pm$ 2.0}
& & 77.0 & 86.4 \\
SFRS \cite{Ge_2020_sfrs} &  4096 & Pitts30k
& \textbf{90.1\footnotesize{$\pm$ 0.3}} & 95.8\footnotesize{$\pm$ 0.2} &
& \textbf{88.7\footnotesize{$\pm$ 0.3}} & 94.2\footnotesize{$\pm$ 0.1} &
& 78.5\footnotesize{$\pm$ 0.8} & 87.3\footnotesize{$\pm$ 0.5} &
& 62.8\footnotesize{$\pm$ 0.8} & 73.0\footnotesize{$\pm$ 0.8} &
& 72.5\footnotesize{$\pm$ 3.6} & 85.4\footnotesize{$\pm$ 3.2}
& & 78.5 & 87.1 \\
SRALNet \cite{Peng_2021_sralNet} \textdagger & 4096 & Pitts30k
& 87.8 & 94.8 & &  -   &  -   & & 72.1 & 83.2 &
&  -   &  -   & &  -   &  -   & &  -   &  -   \\
APPSVR \cite{Peng_2021_appsvr} \textdagger & 4096 & Pitts30k
& 88.8 & 95.6 & & 87.4 & 94.3 & & 77.1 & 85.7 &
&  -   &  -   & &  -   &  -   & &  -   &  -   \\

GeM \cite{Radenovic_2019_gem} &   512 & Pitts30k
& 75.3\footnotesize{$\pm$ 0.2} & 88.4\footnotesize{$\pm$ 0.3} &
& 77.9\footnotesize{$\pm$ 0.4} & 90.5\footnotesize{$\pm$ 0.3} &
& 46.4\footnotesize{$\pm$ 0.9} & 65.3\footnotesize{$\pm$ 0.7} &
& 51.8\footnotesize{$\pm$ 0.9} & 64.4\footnotesize{$\pm$ 0.9} &
& 59.9\footnotesize{$\pm$ 1.6} & 76.3\footnotesize{$\pm$ 2.0}
& & 62.3 & 77.0 \\
GeM \cite{Radenovic_2019_gem} &   512 & MSLS
& 65.3\footnotesize{$\pm$ 1.2} & 81.0\footnotesize{$\pm$ 1.6} &
& 71.6\footnotesize{$\pm$ 2.1} & 85.1\footnotesize{$\pm$ 1.9} &
& 44.9\footnotesize{$\pm$ 1.7} & 62.6\footnotesize{$\pm$ 1.2} &
& 66.7\footnotesize{$\pm$ 0.7} & 78.9\footnotesize{$\pm$ 0.5} &
& 84.6\footnotesize{$\pm$ 1.1} & 93.3\footnotesize{$\pm$ 0.7}
& & 66.6 & 80.2 \\
GeM \cite{Radenovic_2019_gem} &   512 & {\ourD}*
& 64.7\footnotesize{$\pm$ 0.8} & 81.4\footnotesize{$\pm$ 0.8} &
& 67.8\footnotesize{$\pm$ 0.6} & 83.6\footnotesize{$\pm$ 0.7} &
& 37.9\footnotesize{$\pm$ 2.3} & 51.0\footnotesize{$\pm$ 2.1} &
& 46.8\footnotesize{$\pm$ 2.1} & 58.1\footnotesize{$\pm$ 1.2} &
& 68.5\footnotesize{$\pm$ 2.4} & 82.7\footnotesize{$\pm$ 1.8}
& & 57.1 & 71.4 \\
APANet \cite{Zhu_2018_apanet} \textdagger & 512 & Pitts30k
& 83.7 & 92.6 & &  -   &  -   & & 67.0 & 81.0 &
&  -   &  -   & &  -   &  -   & &  -   &  -   \\
\hline
\textbf{{\our} (Ours)} &   512 & {\ourD}
& 89.7\footnotesize{$\pm$ 0.0} & \textbf{96.4\footnotesize{$\pm$ 0.2}} &
& 88.5\footnotesize{$\pm$ 0.2} & \textbf{94.5\footnotesize{$\pm$ 0.0}} &
& \textbf{82.8\footnotesize{$\pm$ 0.9}} & \textbf{90.0\footnotesize{$\pm$ 0.6}} &
& \textbf{79.5\footnotesize{$\pm$ 0.1}} & \textbf{87.2\footnotesize{$\pm$ 0.2}} &
& \textbf{94.3\footnotesize{$\pm$ 0.4}} & 97.4\footnotesize{$\pm$ 0.3}
& & \textbf{87.0} & \textbf{93.1} \\
\bottomrule
\end{tabular}
\end{adjustbox}
\vspace{-0.2cm}
\caption{\textbf{Comparisons of various methods on popular datasets.} Two values of recalls are used as metric (R@1, R@5), with threshold distance for positives of 25 meters. Results rely on a CNN with a VGG-16 backbone, and are averaged over 3 runs with seeds 0, 1 and 2.
* images taken after 2010 are used as queries, and the rest as database.
** images taken after 2015 are used as queries, and the rest as database.
\textdagger these results could not be independently verified given that no public code implementation is available and we were unable to reproduce the results. The results (when available) were taken from the respective papers.
}
\label{tab:comp_all_ds_full_dim}
\vspace{-0.3cm}
\end{table*}

\begin{table}
\begin{adjustbox}{width=\linewidth}
\centering
\begin{tabular}{lccccccccc}
\toprule
\multicolumn{1}{l}{\multirow{2}{*}{Method}} & \multicolumn{1}{c}{\multirow{2}{*}{Desc. dim.}} & \multicolumn{1}{c}{\multirow{2}{*}{Train set}} & \multicolumn{3}{c}{{\ourD} test v1} & & \multicolumn{3}{c}{{\ourD} test v2} \\
\cline{4-6} \cline{8-10}
\multicolumn{3}{c}{}
& R@1  & R@5  & R@10  & & R@1  & R@5  & R@10 \\
\hline
NetVLAD \cite{Arandjelovic_2018_netvlad} & 32768 & Pitts30k
& 40.0 & 48.8 & 52.8 & & 71.1 & 82.4 & 84.8 \\
NetVLAD \cite{Arandjelovic_2018_netvlad} & 32768 & MSLS
& 22.9 & 32.0 & 37.1 & & 48.2 & 64.7 & 69.7 \\
NetVLAD \cite{Arandjelovic_2018_netvlad} & 32768 & {\ourD}*
& 38.3 & 47.4 & 51.7 & & 70.7 & 83.4 & 87.0 \\
NetVLAD \cite{Arandjelovic_2018_netvlad} & 32768 & {\ourD}**
& 28.4 & 38.6 & 43.1 & & 60.9 & 74.6 & 78.6 \\
CRN \cite{Kim_2017_crn}                  & 32768 & Pitts30k
& 45.8 & 56.9 & 60.7 & & 76.4 & 85.3 & 87.8 \\
SARE \cite{Liu_2019_sare}                &  4096 & Pitts30k
& 45.5 & 56.5 & 60.0 & & 78.8 & 87.6 & 90.5 \\
SFRS \cite{Ge_2020_sfrs}                 &  4096 & Pitts30k
& 51.2 & 62.2 & 66.6 & & 83.1 & 90.5 & 93.5 \\
GeM \cite{Radenovic_2019_gem}            &   512 & Pitts30k
& 21.7 & 30.3 & 34.4 & & 43.1 & 63.7 & 69.2 \\
GeM \cite{Radenovic_2019_gem}            &   512 & MSLS
&  8.1 & 15.6 & 20.2 & & 29.3 & 46.3 & 53.8 \\
GeM \cite{Radenovic_2019_gem}           &   512 & {\ourD}*
&  9.8 & 17.6 & 21.2 & & 34.8 & 55.5 & 63.0 \\
\hline
\textbf{{\our} (Ours)} & 512 & {\ourD}
& \textbf{64.7} & \textbf{73.3} & \textbf{76.6} & & \textbf{83.4} & \textbf{91.6} & \textbf{94.1} \\
\bottomrule
\end{tabular}
\end{adjustbox}
\vspace{-0.2cm}
\caption{\textbf{Comparisons on {\ourD} test v1 and {\ourD} test v2.}}
\label{tab:comp_sf_xl_full_dim}
\vspace{-0.3cm}
\end{table}

\myparagraph{Discussion of results.}
Results are reported in \cref{tab:comp_all_ds_full_dim} and \cref{tab:comp_sf_xl_full_dim}, and all rely on VGG-16-based architectures.
The results can be summarized in a few points: \\
- {\our} achieves best results on average, outperforming the second-best method by 8.5\% of R@1 averaged over five datasets. \\
- {\our} shows strong robustness to datasets coming from other sources. On the other hand, other methods either perform well on datasets with StreetView-sourced database (\ie, Pitts30k, Pitts250k, Tokyo 24/7) or on datasets with frontal view images (\ie, MSLS and St Lucia).
For example, the best performing method on MSLS and St Lucia, namely NetVLAD trained on MSLS, falls short of 10.0\% of recall@1 on Pitts250k w.r.t. {\our};
similarly, the best performing on Pitts30k, namely SFRS trained on Pitts30k, is outperformed by {\our} by 21.8\% on St Lucia. \\
- While other methods fail to reach competitive results with compact descriptors, we find that {\our} reaches state-of-the-art with 512-D outputs, paving the way for scalable, robust, and efficient real-world applications. 
In \cref{sec:comparisons_with_pca}, we provide additional results by applying PCA to high-dimensionality descriptors. \\
- Methods besides {\our} do not benefit from training on {\ourD}. This is likely due to the partial mining \cite{Warburg_2020_msls} (given that standard mining \cite{Arandjelovic_2018_netvlad} is unfeasible on a large scale). \\
- For methods other than {\our}, the way database and queries are partitioned in the training set makes a big difference: training on {\ourD}* versus {\ourD}**, which use different database/queries partitions, achieve quite different results.
This is not a problem for {\our} since it does not need a database/queries split for training.

\myparagraph{Fairness of comparisons.}
While it can be argued that comparisons in \cref{tab:comp_all_ds_full_dim} and \cref{tab:comp_sf_xl_full_dim} are not fair, given that most other methods are trained on Pitts30k, and {\our} on a much larger dataset, we want to point out that
i) {\our} cannot be trained on Pitts30k nor MSLS (given that they lack heading labels),  and 
ii) training previous state-of-the-art methods \cite{Liu_2019_sare, Ge_2020_sfrs} on {\ourD} would require a huge amount of resources, and it is practically unfeasible: 
computing the cache (which contains all database and queries descriptors) involves a forward pass on 40M images, with a memory requirement of 5TB, and it would have to be computed periodically every 1000 iterations.

With these considerations in mind, we were able to use the partial mining method from \cite{Warburg_2020_msls} to train GeM and NetVLAD, which have already been shown to perform competitively with such mining \cite{Warburg_2020_msls}, whereas more modern techniques, such as SARE \cite{Liu_2019_sare} or SFRS \cite{Ge_2020_sfrs}, cannot be used with partial mining without requiring significant changes to their algorithm.
However, the fact that GeM and NetVLAD do not benefit from training on the larger scale of SF-XL proves that our superior results do not come simply from the dataset size and that the proposed algorithm is needed to succeed in city-wide geo-localization.

Moreover, following \cite{Arandjelovic_2018_netvlad, Kim_2017_crn, Liu_2019_sare, Ge_2020_sfrs, Peng_2021_appsvr, Peng_2021_sralNet, Yu_2020_SPEVlad} we did not use as comparisons re-ranking based methods \cite{Hausler_2021_patch_netvlad, Cao_2020_delg, Berton_2021_geowarp, Fuwen_2021_reranking_transformers}, given that they perform an extra post-processing step, while methods like {\our} simply rely on a more efficient retrieval through a nearest neighbor search.


\subsection{Computational efficiency}

\myparagraph{Train-time memory footprint.}
Compared to previous methods that use a triplet loss and require to keep the descriptors for each image \cite{Arandjelovic_2018_netvlad, Liu_2019_sare, Ge_2020_sfrs} (or a large number of them \cite{Warburg_2020_msls}) in a cache, {\our} circumvents this by posing the problem as a classification task without mining/caching.
This effectively means that it can scale to very large datasets, whereas previous works \cite{Arandjelovic_2018_netvlad, Liu_2019_sare, Ge_2020_sfrs, Yu_2020_SPEVlad, Peng_2021_appsvr, Peng_2021_sralNet} would require massive amounts of memory to train on large-scale datasets.
Note that a cache for a large dataset could be extremely memory demanding: NetVLAD descriptors for an image weigh about $32\textrm{k} \cdot 4\textrm{B}=128\textrm{KB}$, while a JPEG image of dimension $480\times640$ is about 70KB; therefore, the cache for a whole dataset is almost twice as heavy as the dataset itself.

\myparagraph{GPU requirements.}
While previous state of the art (\ie, SFRS \cite{Ge_2020_sfrs}) requires four 11GB GPUs to train, {\our} is much lighter.
Using the same backbone (a VGG-16 \cite{Simonyan_2015_vgg}), we only require 7.5 GB on a single GPU to obtain the results shown in \cref{tab:comp_all_ds_full_dim} and \cref{tab:comp_sf_xl_full_dim}.

\myparagraph{Descriptors dimension.}
Since the advent of the groundbreaking NetVLAD paper \cite{Arandjelovic_2018_netvlad}, state of the art relied on the NetVLAD layer \cite{Arandjelovic_2018_netvlad, Kim_2017_crn, Liu_2019_sare, Ge_2020_sfrs}, which produces high-dimensional descriptors.
Considering that the compactness of the descriptors is perhaps one of the most important factors when choosing an algorithm in the real world (given that for efficient retrieval, all database descriptors should be kept in memory), we produce much more compact vectors w.r.t. previous methods.
For instance, while \cite{Arandjelovic_2018_netvlad, Kim_2017_crn} use the full NetVLAD dimension of 32k, and \cite{Liu_2019_sare, Ge_2020_sfrs} reduce it to 4k, {\our} achieves SOTA results with a dimension of just 512.
Moreover, in \cref{sec:backbones} and \cref{sec:further_backbones}, we investigate how results depend on the dimensionality of the descriptors (and the underlying backbones), showing that {\our} with a ResNet-101 and 128-D descriptors outperforms previous SOTA.

\myparagraph{Training and testing speed.}
The network takes roughly one day to train on {\ourD}, similarly to SFRS's \cite{Ge_2020_sfrs} training time on Pitts30k.
At test time, with a VGG-16-based model, a V100 GPU is able to extract descriptors for 80 images per second, although we find that the speed largely depends on the backbone and is very similar among the methods of \cref{tab:comp_all_ds_full_dim}, given that all rely on a VGG-16.
However, for large-scale datasets, the time required for the k-Nearest Neighbors search dwarfs the extraction time (considering a real-world scenario where database descriptors are extracted offline).
Given that exhaustive kNN's execution time linearly depends on the dimensionality of the descriptors, our 512-D network is 8 times faster than 4096-D SFRS for inference on large datasets and 64 times faster than 32k-D NetVLAD.

\begin{table}
\centering
\begin{adjustbox}{width=0.6\linewidth}
\begin{tabular}{ccccc}
\toprule
\multirow{2}{*}{\begin{tabular}[c]{@{}c@{}}M\\(meters)\end{tabular}} &
\multirow{2}{*}{\begin{tabular}[c]{@{}c@{}}$\alpha$\\(degrees °)\end{tabular}} &
\multicolumn{1}{c}{\multirow{2}{*}{N}} &
\multicolumn{1}{c}{\multirow{2}{*}{L}} &
\multicolumn{1}{c}{\multirow{2}{*}{SF-XL val set}} \\ \\
\hline
10 & 360 & 1 & 1 & 85.8 \\
10 & 360 & 5 & 1 & 88.5 \\
10 &  30 & 5 & 1 & 88.2 \\
10 &  30 & 1 & 2 & 77.1 \\
10 &  30 & 5 & 2 & \textbf{90.9} \\
\bottomrule
\end{tabular}
\end{adjustbox}
\caption{\textbf{Ablation}. Recall@1 with varying values of hyperparameters $M$, $\alpha$, $N$ and $L$.
The bottom row represents {\our}, and the other rows represent {\our} without some components:
for example using $\alpha = 360\degree$ means that heading labels are not used (\ie all images within any given cell belong to the same class); similarly using $N=1$ and $L=1$ means that all classes belong to the same group (and only one group exists, resulting in no group separation). A more thorough ablation is in the {\supplementary}.}
\label{tab:ablation}
\vspace{-0.3cm}
\end{table}

\subsection{Ablation}
\label{sec:ablation}
\myparagraph{Hyperparameters.}
To better understand how the choices made in \cref{sec:method} impact the results, we perform a series of experiments, reported in \cref{tab:ablation}.
The results show that any naive split of the dataset into classes fails to achieve competitive results.
The fact that the last row (which represents the only experiment computed without any intra-group adjacent classes) achieves considerably better results than the other rows clearly proves the benefit of using {\our} groups.

\begin{figure}
    \centering
    \includegraphics[width=0.99\linewidth]{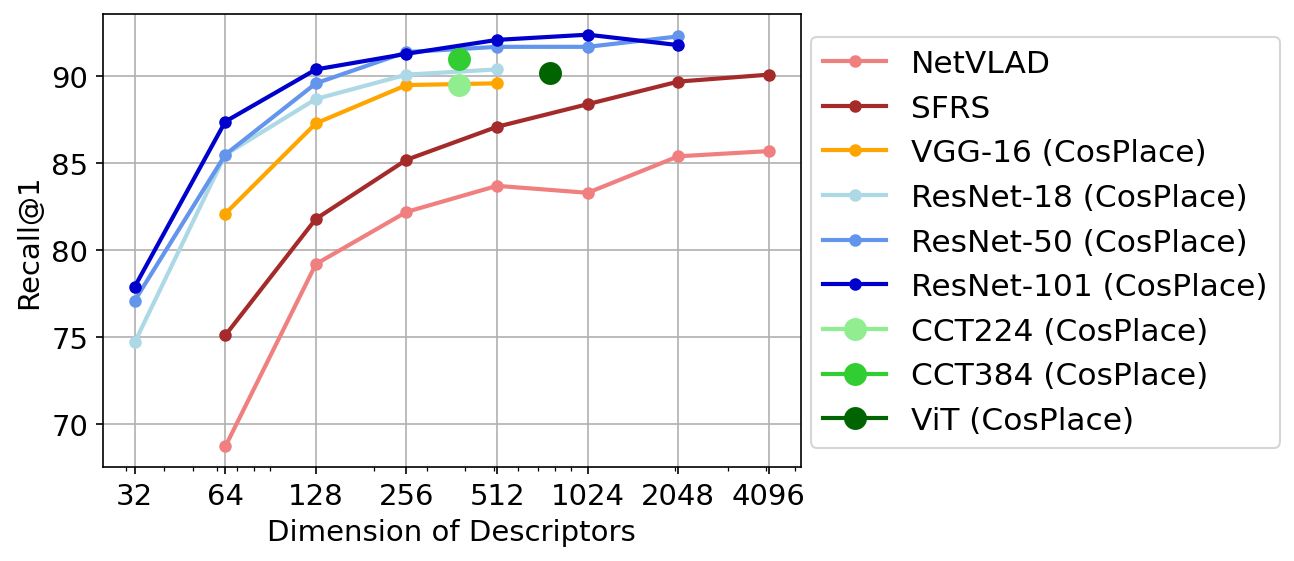}
    \vspace{-0.2cm}
    \caption{Results on Pitts250k of \textbf{{\our} using different backbones and dimensionality of descriptors}, compared with SFRS and NetVLAD trained on Pitts30k.
    }
    \label{fig:backbones_main}
    \vspace{-0.3cm}
\end{figure}

\myparagraph{Backbones and descriptors dimensionality.}
\label{sec:backbones}
In \cref{fig:backbones_main}, we explore the use of different backbones, such as ResNets \cite{He_2016_resnet} and Transformers \cite{Dosovitskiy_2021_vit, Hassani_2021_cct}.
We find that {\our} performs well with a variety of backbones, with no need to apply any changes to hyperparameters or layers.
This is in contrast with NetVLAD-based architectures (e.g. ResNets followed by NetVLAD perform best when the backbone is cropped at the fourth residual layer instead of the last one \cite{Berton_2022_benchmark}).
Given that, when using {\our}, ResNets outperform the VGG-16 while generally being faster and requiring less memory, we believe that future works should move away from the outdated use of VGG-16 backbones towards more efficient ResNets.
More details and results on different backbones and descriptors dimensionality are in \cref{sec:further_backbones}.


\subsection{Limitations}
\myparagraph{Heading labels.}
Unlike previous VG methods, which solely rely on GPS/UTM coordinates, we also take advantage of heading labels.
This is a drawback of {\our}, although we argue that heading labels are really inexpensive, and just like GPS coordinates, they can be collected simply through a sensor, without requiring any manual annotation whatsoever.
However, some commonly used datasets such as Pitts30k and MSLS do not provide heading labels for their images, making {\our} not trainable on such datasets.

\begin{table}
\centering
\begin{adjustbox}{width=0.8\linewidth}
\begin{tabular}{lccccccccccccccccccc}
\toprule
\multicolumn{1}{c}{\multirow{2}{*}{\# images}} &
\multicolumn{1}{c}{\multirow{2}{*}{Pitts250k}} &
\multicolumn{1}{c}{\multirow{2}{*}{Pitts30k}} &
\multicolumn{1}{c}{\multirow{2}{*}{Tokyo 24/7}} &
\multicolumn{1}{c}{\multirow{2}{*}{MSLS}} &
\multicolumn{1}{c}{\multirow{2}{*}{St Lucia}} \\ \\
\hline
5.6M & \textbf{90.4} & \textbf{89.5} & \textbf{81.6} & \textbf{81.8} & \textbf{98.8} \\
120k & 86.0 & 85.9 & 67.9 & 73.3 & 96.7 \\
12k  & 82.1 & 82.9 & 61.3 & 66.2 & 91.2 \\
1.2k & 73.3 & 75.9 & 45.7 & 55.0 & 85.2 \\
\bottomrule
\end{tabular}
\end{adjustbox}
\caption{\textbf{Limited downward scalability.} The first column represents the number of images which are effectively used at training time (sampled from {\ourD}), and the other columns show the recall@1 on multiple datasets.}
\label{tab:smaller_training_dataset}
\vspace{-0.3cm}
\end{table}

\myparagraph{Limited downward scalability.}
\label{sec:smaller_training_dataset}
Although {\our} has been designed for a large-scale dataset, in this paragraph we investigate how the use of a smaller training dataset influences the results.
In \cref{tab:smaller_training_dataset}, we show how recalls vary for logarithmically decreasing sizes of the training dataset.
We find that {\our} needs a large number of training images to reach SOTA performance, and therefore it is not suited to be used for training on small datasets.

\section{Conclusions}
In this work, we study the task of visual geo-localization (VG) in large-scale applications. Using the newly proposed San Francisco eXtra Large ({\ourD}) dataset, we find that current methods based on contrastive learning are hardly scalable to train on large quantities of data.
To address this problem, we propose a new method, {\our}, which allows us to train on large quantities of data efficiently.
We demonstrate that very simple architectures trained on {\ourD} using {\our} can surpass the current state of the art while using an 8x smaller descriptor.
Most importantly, {\our} is remarkably simple, needs much less training resources than the contrastive-based approaches, and generalizes extremely well to other domains.
Although {\our} has some limitations, \ie, it is not suited for training on small datasets nor datasets without orientation labels, it paves the way for a new strategy to tackle VG in large-scale applications.

\noindent\textbf{Acknowledgements.}
\small{We acknowledge the CINECA award under the ISCRA initiative, for the availability of high performance computing resources.
This work was supported by CINI.}

{\small
\bibliographystyle{ieee_fullname}
\bibliography{egbib}
}

\appendix

\section*{Appendix}
In the following, \cref{sec:further_information_on_sf_xl} provides additional information on San Francisco eXtra Large ({\ourD}),
\cref{sec:further_ablations} presents a thorough ablation over all {\our} hyperparameters,
and \cref{sec:further_implementation_details} provides further implementation details on {\our}.
Finally, in \cref{sec:exploratory_experiments}, we provide a large set of extra results, comprising
experiments on changing the backbones and the descriptors dimensionality, and further comparisons with other methods.

\section{Further information on {\ourD}}
\label{sec:further_information_on_sf_xl}

\begin{figure}
    \centering
    \includegraphics[width=0.8\linewidth]{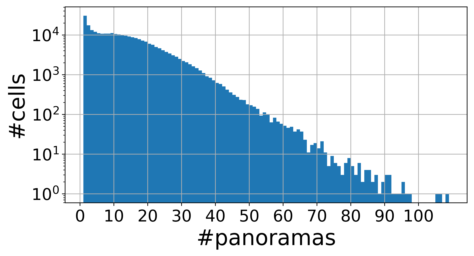}
    \vspace{-0.3cm}
    \caption{\textbf{Histogram showing how many cells contain a given number of panorama.} We can see that cells with only one panorama (which are discarded at train time as explained in \cref{sec:further_implementation_details}) are the most common. Note that the y axis is in logarithmic scale. The side of each cell (i.e. the hyperparameter $M$) is $M=10$ meters, as in our final experiments.}
    \label{fig:num_cells_with_pano}
\end{figure}

\begin{figure}
    \centering
    \includegraphics[width=0.8\linewidth]{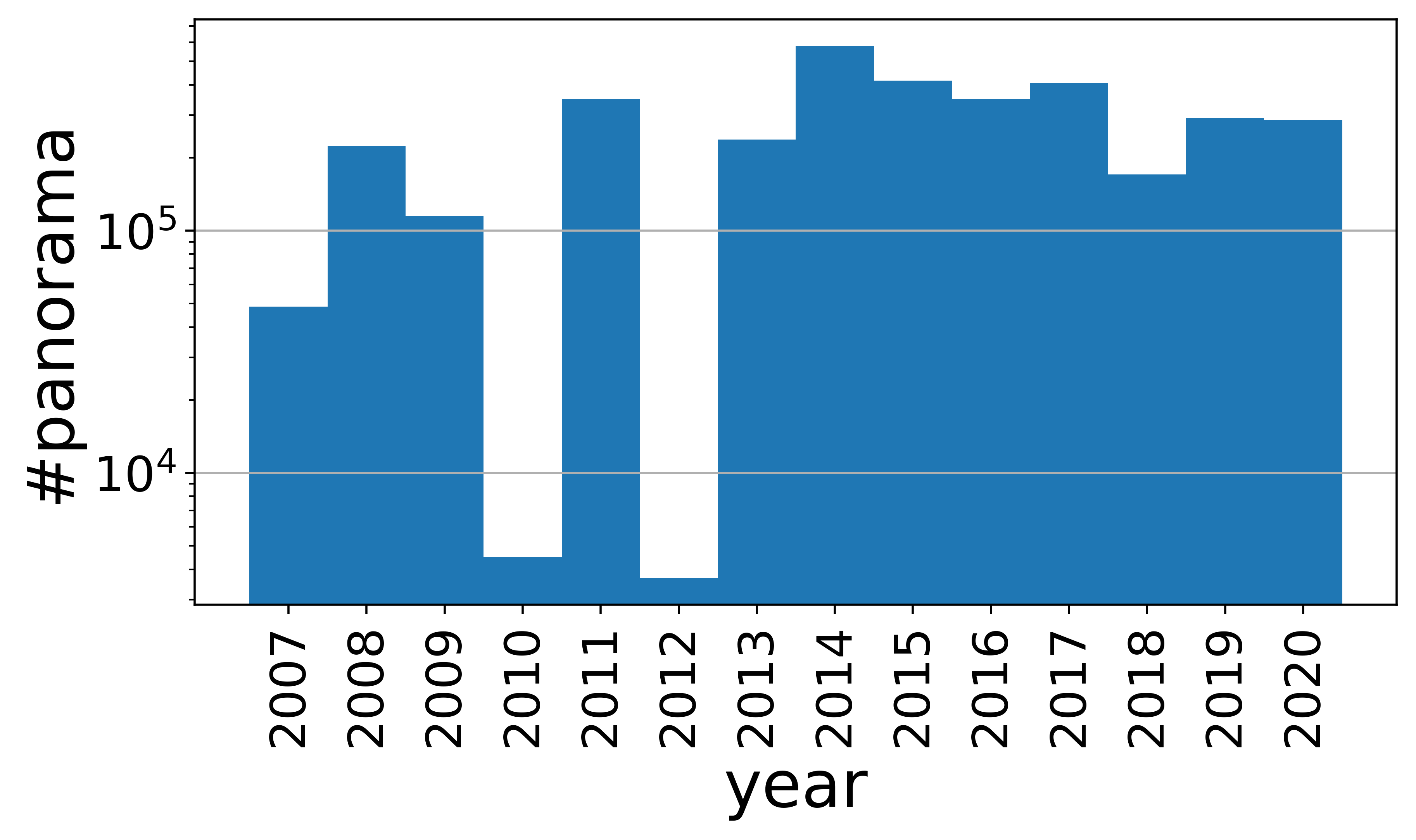}
    \vspace{-0.2cm}
    \caption{\textbf{Number of 360\degree panorama of {\ourD} for each given year.}}
    \label{fig:panos_per_year}
    \vspace{-0.2cm}
\end{figure}

\begin{figure}
    \centering
    \includegraphics[width=0.8\linewidth]{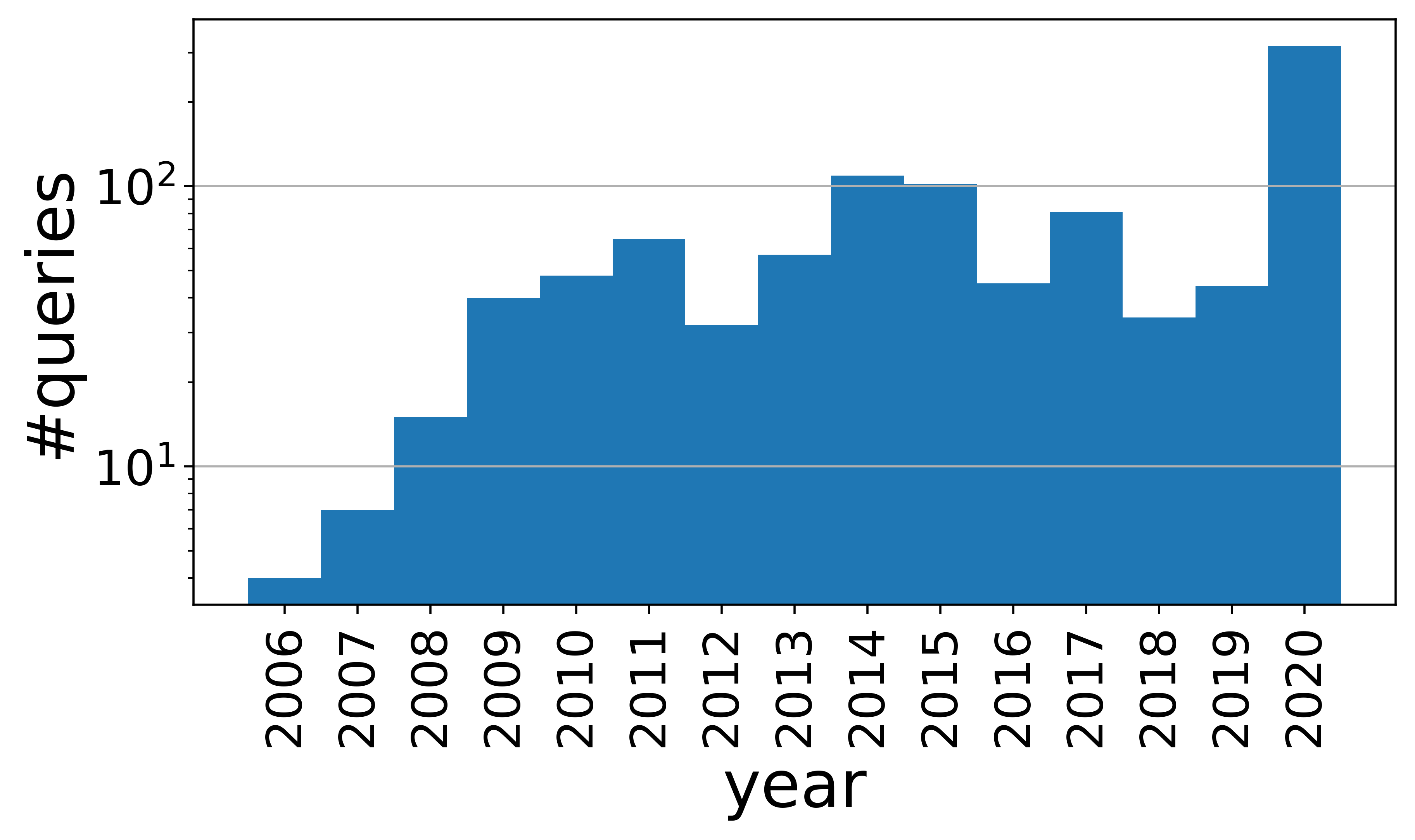}
    \vspace{-0.2cm}
    \caption{\textbf{Number of queries from {\ourD} test v1 for each given year.}}
    \vspace{-0.5cm}
    \label{fig:queries_per_year}
\end{figure}

\myparagraph{General information.}
In \cref{fig:num_cells_with_pano} we show the density of the training set (\ie number of panoramas within each cell), in \cref{fig:panos_per_year} is its temporal distribution,
and in \cref{fig:queries_per_year} we show the temporal variability of {\ourD} test v1's queries.
The StreetView images composing the train set, val set and test database, are $512 \times 512$ images cropped from 360{\degree} panoramas.

\begin{figure*}[t!]
    \centering
    \begin{subfigure}{\textwidth}
        \includegraphics[width=\textwidth]{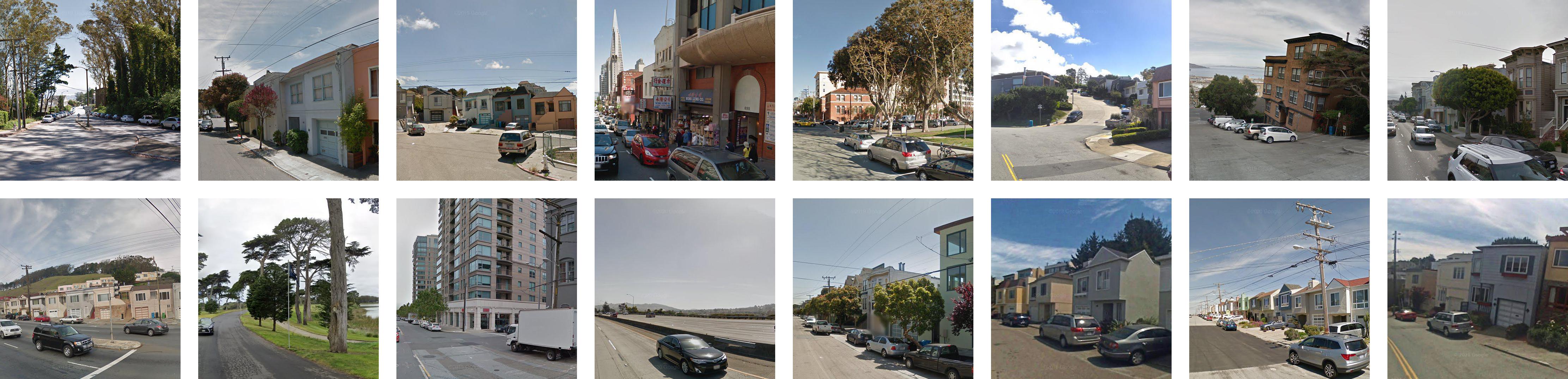}
    \end{subfigure}
    \vspace{5pt}
    \par\noindent\rule{\textwidth}{0.5pt}
    \rule{\textwidth}{0.pt}
    \vspace{6pt}
    \begin{subfigure}{\textwidth}
        \includegraphics[width=\textwidth]{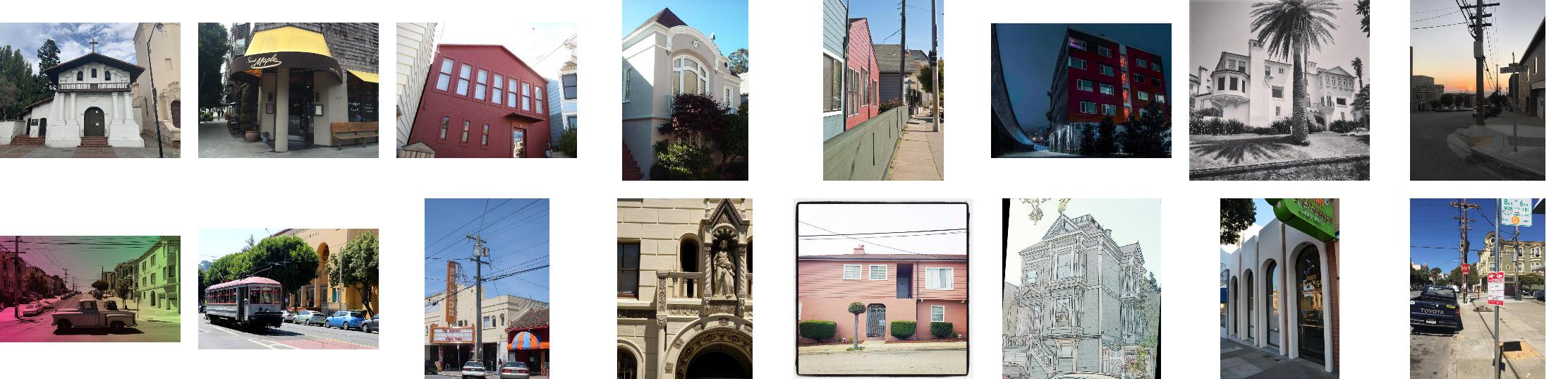}
    \end{subfigure}
    \vspace{3pt}
    \par\noindent\rule{\textwidth}{0.5pt}
    \rule{\textwidth}{0.pt}
    \begin{subfigure}{\textwidth}
        \includegraphics[width=\textwidth]{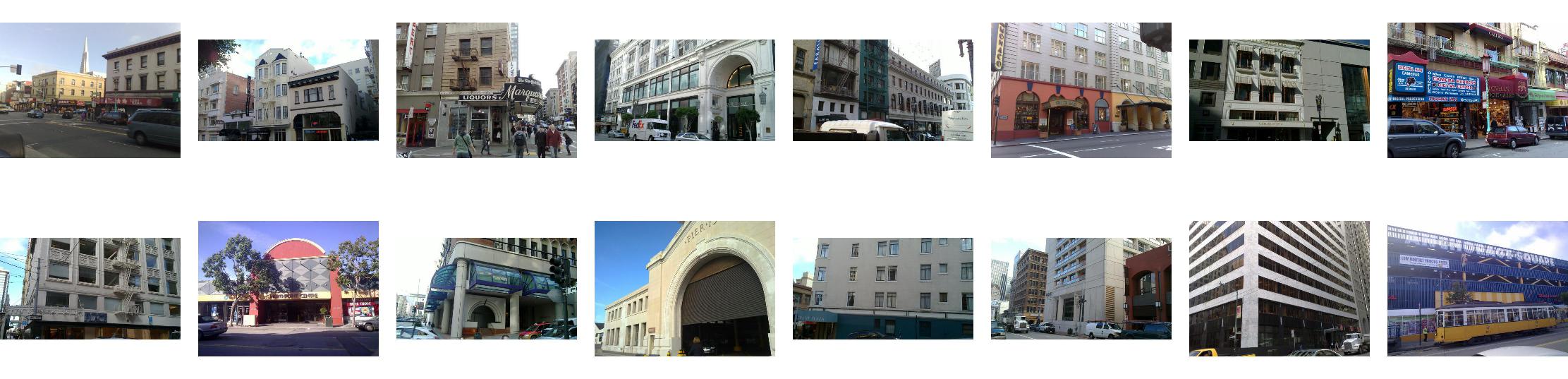}
    \end{subfigure}
    \caption{\textbf{Examples from {\ourD}}. The first two rows of images are from the train set, the next two from the queries of {\ourD} test v1, and the last two rows from the queries of {\ourD} test v2.
    }
    \label{fig:examples}
\end{figure*}

\myparagraph{{\ourD} test v1.}
While the database from {\ourD} test v1 is very homogeneous, given that StreetView images are all taken at daytime with the same camera and good weather, the queries present large degrees of domain changes: there are night images, grayscale, with heavy changes in viewpoint and occlusions.
Coming from the crowd-sourced platform Flickr, these queries are collected by a large number of users, also ensuring variety in the typologies of cameras.
We resized these images so that their shorter side is 480 pixels.
In \cref{fig:examples}, we show more examples of queries, besides the ones shown in Fig. 2 of the main paper.

\myparagraph{{\ourD} test v2.}
While the database of {\ourD} test v2 is the same as {\ourD} test v1, the two sets use different queries.
With the advantage of having 6 DoF labels, {\ourD} test v2 can also be used for pose estimation.
The downside of this set is the homogeneity among its images, given that almost all are taken during sunny days, with clear views and without heavy occlusions.
Some examples of the queries are shown in \cref{fig:examples}.

\section{Experiments}
\subsection{Further ablations}
\label{sec:further_ablations}
In this section, we provide further results obtained by changing the hyperparameters of {\our} to better understand their correlations to the final results.
\begin{figure*}
    \centering
    \begin{minipage}{.19\textwidth}
        \includegraphics[width=\textwidth]{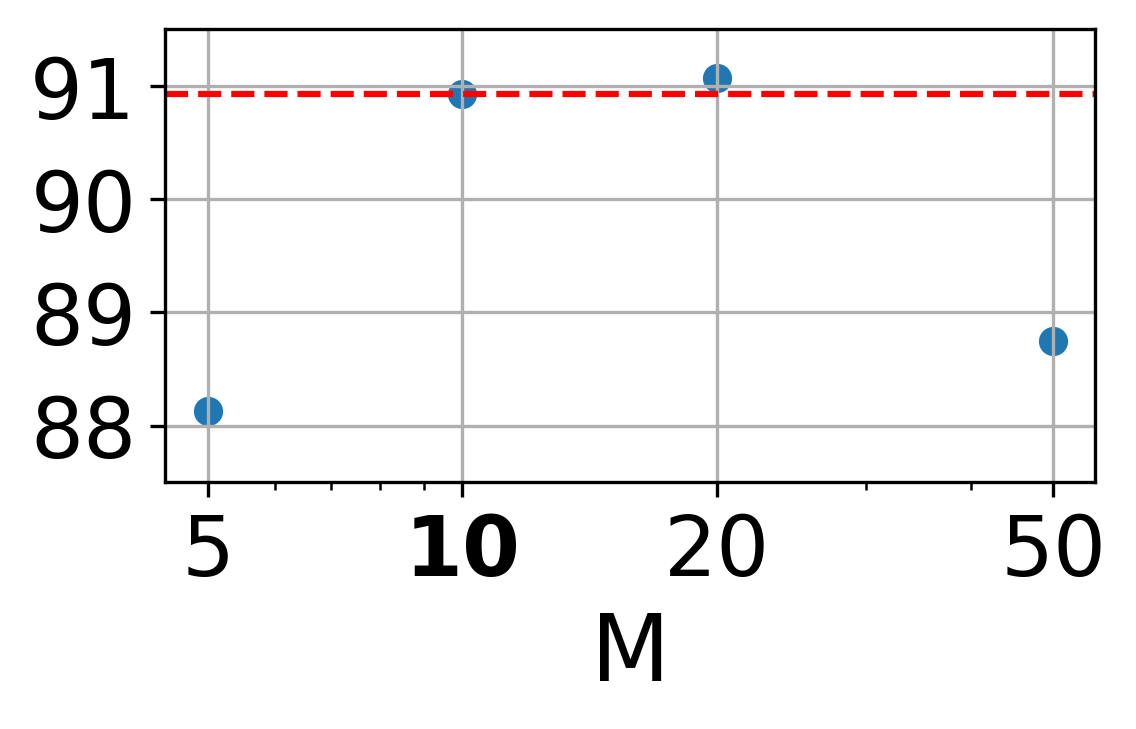}
    \end{minipage}
    \begin{minipage}{.19\textwidth}
        \includegraphics[width=\textwidth]{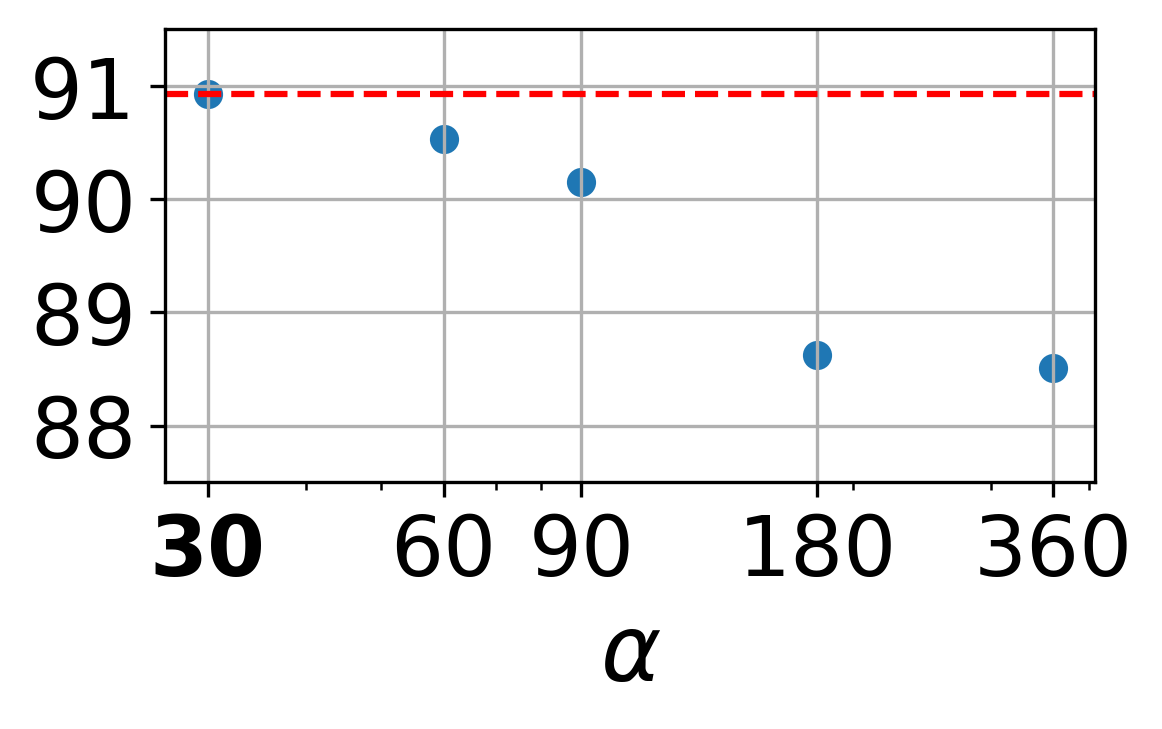}
    \end{minipage}
    \begin{minipage}{.19\textwidth}
        \includegraphics[width=\textwidth]{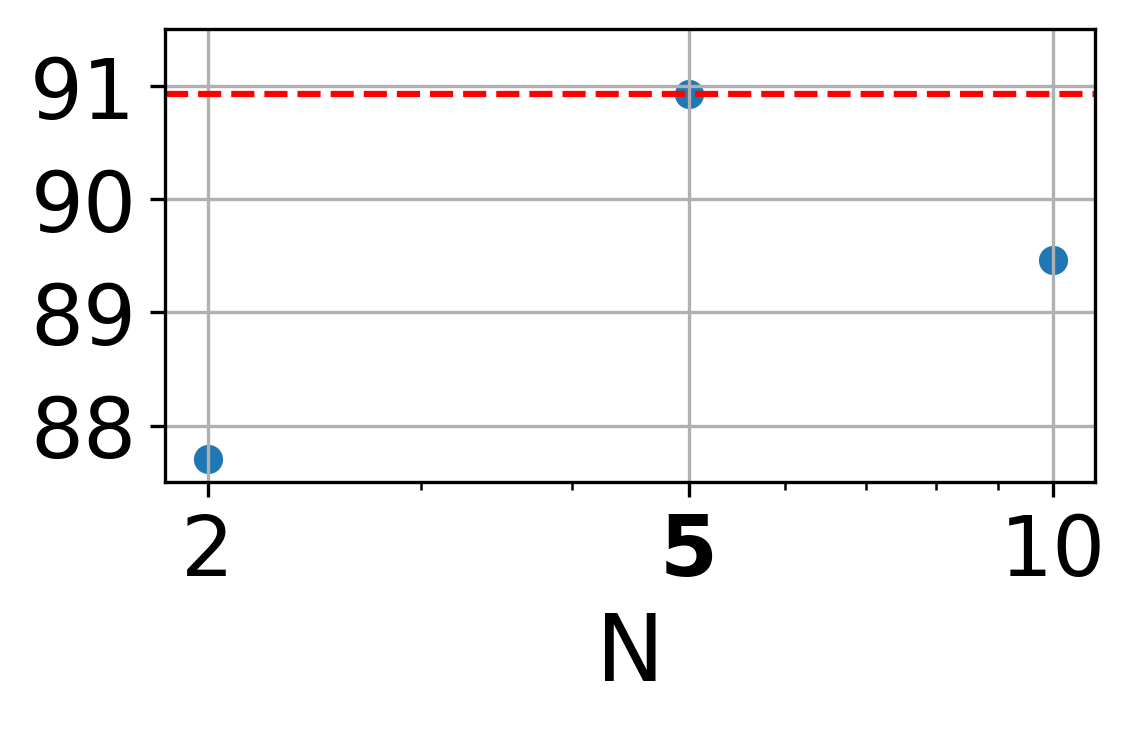}
    \end{minipage}
    \begin{minipage}{.19\textwidth}
        \includegraphics[width=\textwidth]{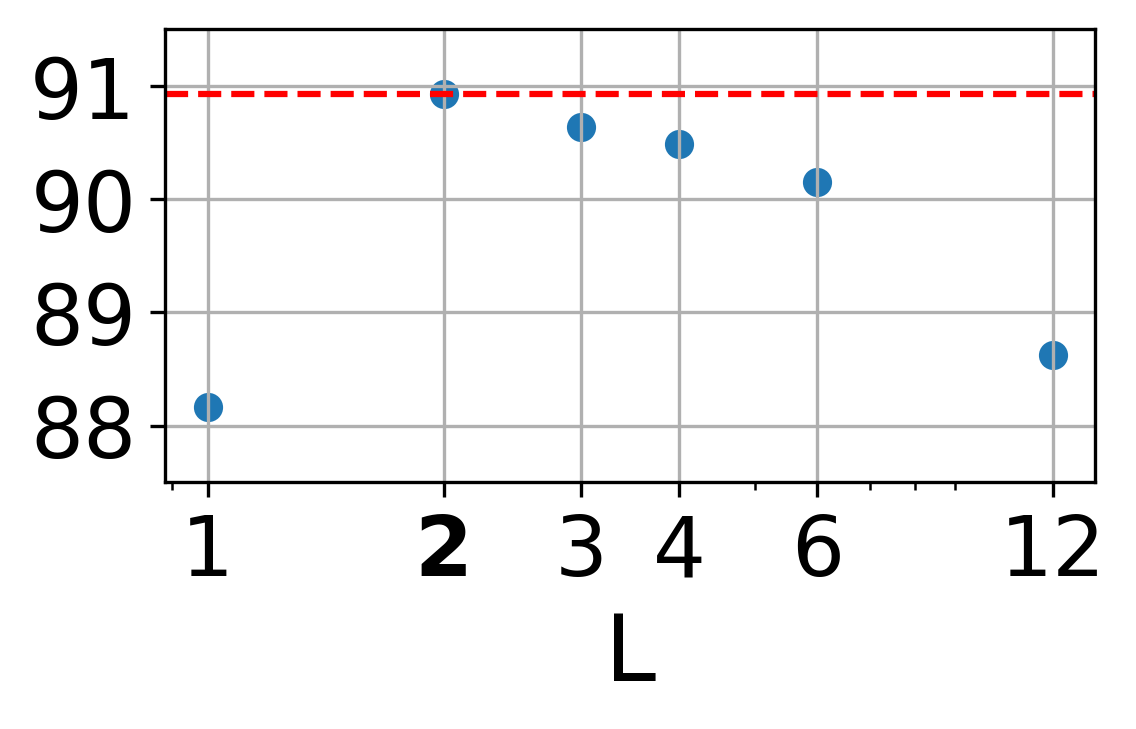}
    \end{minipage}
    \begin{minipage}{.19\textwidth}
        \includegraphics[width=\textwidth]{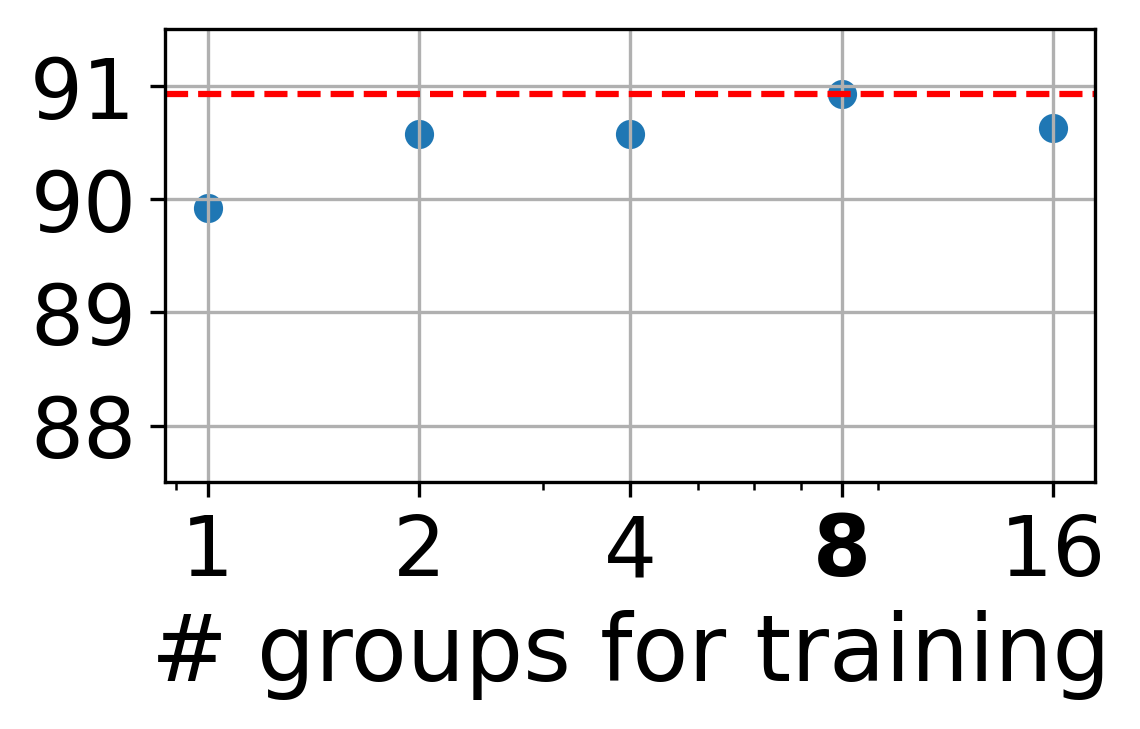}
    \end{minipage}
    \caption{\textbf{Full ablation on each hyperparameter.} On the x axis are values for the hyperparameters, and on the y axis their respective recall@1 on the SF-XL val set, computed with a ResNet-18. Values in bold are the chosen ones for all experiments besides ablations, and the red line represents their recall@1.}
    \label{fig:ablation_full}
\end{figure*}

In \cref{fig:ablation_full} we report an extensive ablation obtained by changing the parameters used to split the dataset into groups and classes, namely $M$, $\alpha$, $N$ and $L$, as well as using a different number of groups for training.
Among other results, we see in the rightmost plot that using just a single group for training the model leads to a drop in recall@1 of just 1\%, and that the optimal results are achieved using 8 of the 50 groups.

To better understand the importance that the GeM pooling \cite{Radenovic_2019_gem} has within the architecture used for {\our}, we provide a set of experiments by replacing it with the average or max pooling in \cref{tab:poolings}.
From the table, we can see that {\our} would outperform the previous state-of-the-art even with a standard architecture used for classification, made of a CNN backbone, a max pooling, and a fully connected layer.

\begin{table}
\begin{adjustbox}{width=\columnwidth}
\centering
\begin{tabular}{lccccc}
\toprule
Pooling & Pitts250k & Pitts30k & Tokyo 24/7 & MSLS & St Lucia \\
\hline
Average & 88.5 & 87.6 & 73.7 & 78.5 & 98.7 \\
Max     & \textbf{90.8} & 89.3 & 78.1 & 80.5 & 98.7 \\
GeM     & 90.4 & \textbf{89.5} & \textbf{81.6} & \textbf{81.8} & \textbf{98.8} \\
\bottomrule
\end{tabular}
\end{adjustbox}
\caption{\textbf{Ablation over different pooling layers.} This table shows results obtained by replacing the GeM layer with a max or average pooling. Results refer to the recall@1 obtained with a ResNet-18.}
\label{tab:poolings}
\end{table}

\subsection{Further implementation details}
\label{sec:further_implementation_details}
Regarding {\our} training, to ensure that each class is well represented, only cells with at least 10 panoramas are considered for training, effectively discarding about 15\% of the images.
The hyperparameters of $M=10$, $\alpha=30$, $N=5$, and $L=2$ lead to the creation of 50 groups, where each group ends up with roughly 35k classes, and each class contains on average 19.8 images.
As explained in the main paper, we only train on 8 (out of 50) groups, which together contain roughly 5.6M images.
Note that the total size of the {\ourD} training set is 41.2M (\ie, we only use 13.6\% of the images), meaning that train-time scalability is a factor that can still be vastly improved in future works.

We use the Adam optimizer \cite{Kingma_2014_adam} with a learning rate of 0.00001, and a batch size of 32 images.
We use color jittering as in \cite{Ge_2020_sfrs}. For results to be fair with \cite{Ge_2020_sfrs}, which uses a smart region cropping method, we also employ random cropping.
Finally, the margin of the cosFace loss is set to 0.40.

\myparagraph{Number of hyperparameters.}
Although {\our} introduces a considerable amount of hyperparameters, we also note that there is no more need for many other ones used in previous state-of-the-art methods \cite{Arandjelovic_2018_netvlad, Liu_2019_sare, Ge_2020_sfrs}, such as the number of negatives per query (usually set to 10), refresh rate of the cache (1000), pool size of randomly sampled negatives (1000), threshold distance for train-time potential positives (10 meters) and the number of cluster in NetVLAD layer (64).
Moreover, the intuitive meaning of the hyperparameters in {\our} in comparison to the less obvious mining hyperparameters makes it easier to set them using common sense: for example, it is clear that a small $M$ (or $\alpha$) leads to little intra-class spatial variations, while a large $M$ (or $\alpha$) may cause two images of the same class to be too different; similarly, using a small value for $N$ leads to a higher similarity between inter-class (but same group) images, while using a very high $N$ leads to classes being very geographically spread out, which can be a problem with smaller datasets (because groups would have few classes).


\subsection{Exploratory experiments}
\label{sec:exploratory_experiments}
\subsubsection{Further results on backbones and descriptors dimensionality.}
\label{sec:further_backbones}
Given that previous methods (as recent as 2021) in Visual geo-localization rely on relatively old VGG-16 \cite{Simonyan_2015_vgg} or AlexNet \cite{Krizhevsky_2012_alexnet} backbones \cite{Arandjelovic_2018_netvlad, Kim_2017_crn, Liu_2019_sare, Ge_2020_sfrs, Zhu_2018_apanet, Peng_2021_sralNet, Peng_2021_appsvr}, we believe that this is widening an already large gap between research and real-world applications, where one would want to obtain the best possible results with the lowest computational complexity.
To narrow such a gap, we investigate how the use of more recent backbones can enhance {\our} and lead to better results, smaller descriptors, and faster computation.
To this end, we train {\our} using a number of backbones, namely VGG-16 \cite{Simonyan_2015_vgg}, ResNet-18, ResNet-50, ResNet-101 \cite{He_2016_resnet}, ViT \cite{Dosovitskiy_2021_vit}, CCT224 and CCT384 \cite{Hassani_2021_cct}.
All CNN backbones (\ie, VGG-16 and ResNets) are followed by a GeM pooling \cite{Radenovic_2019_gem} and a fully connected layer,
Moreover, we experiment with various powers of 2 (from 32-D to 2048-D) as output dimension.
Regarding transformers-based neural networks, we use the 384-D SeqPool output of CCT as descriptors, and for ViT, we obtain the 768-D output by feeding the CLS token to a multi-layer perceptron with tanh, following its original implementation \cite{Dosovitskiy_2021_vit}.
Preliminary results showed that directly using ViT's CLS token led to lower recalls.

\begin{figure*}[t!]
    \centering
    \begin{minipage}{.32\textwidth}
        \begin{subfigure}{\textwidth}
            \includegraphics[width=\textwidth]{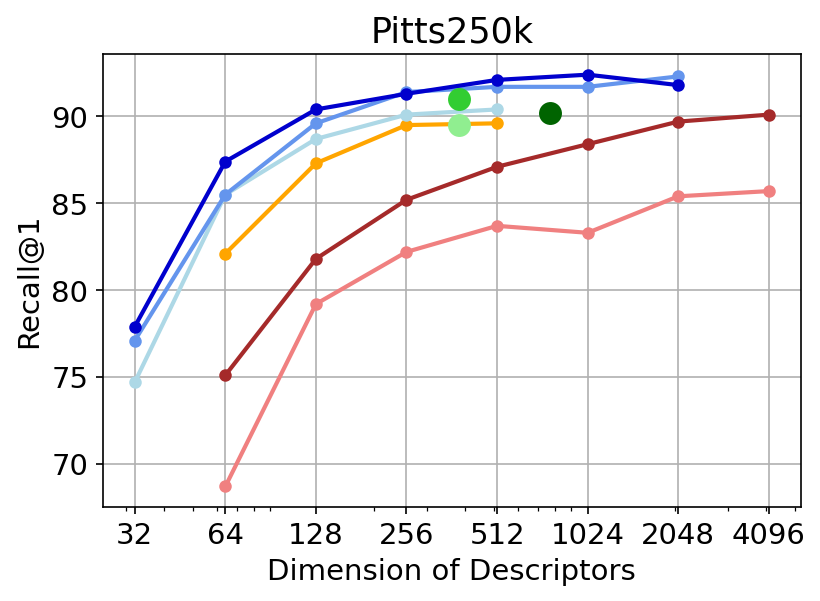}
        \end{subfigure}
        \begin{subfigure}{\textwidth}
            \includegraphics[width=\textwidth]{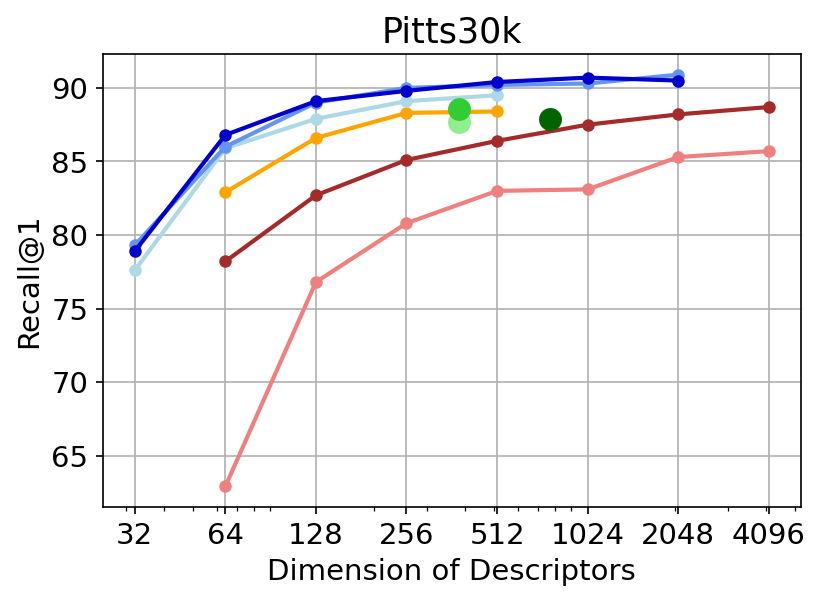}
        \end{subfigure}
    \end{minipage}
    \begin{minipage}{.32\textwidth}
        \begin{subfigure}{\textwidth}
            \includegraphics[width=\textwidth]{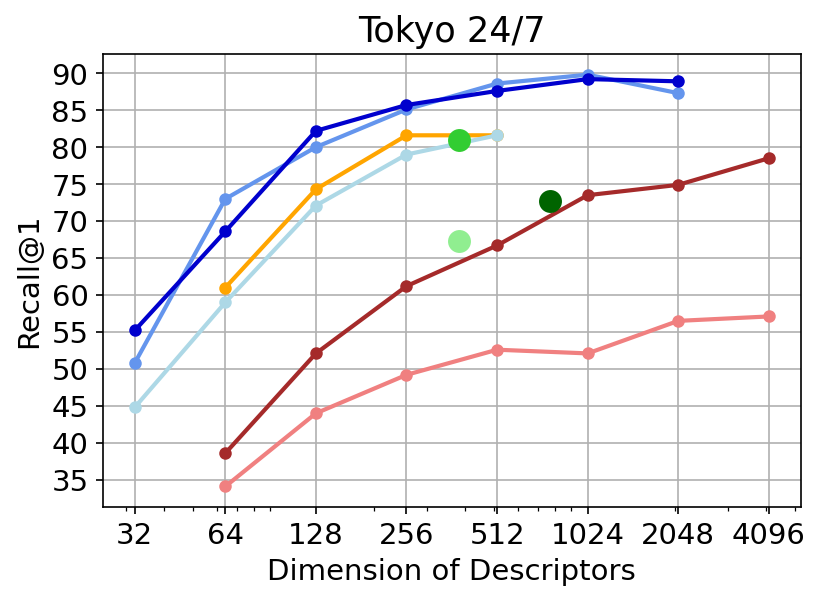}
        \end{subfigure}
        \begin{subfigure}{\textwidth}
            \includegraphics[width=\textwidth]{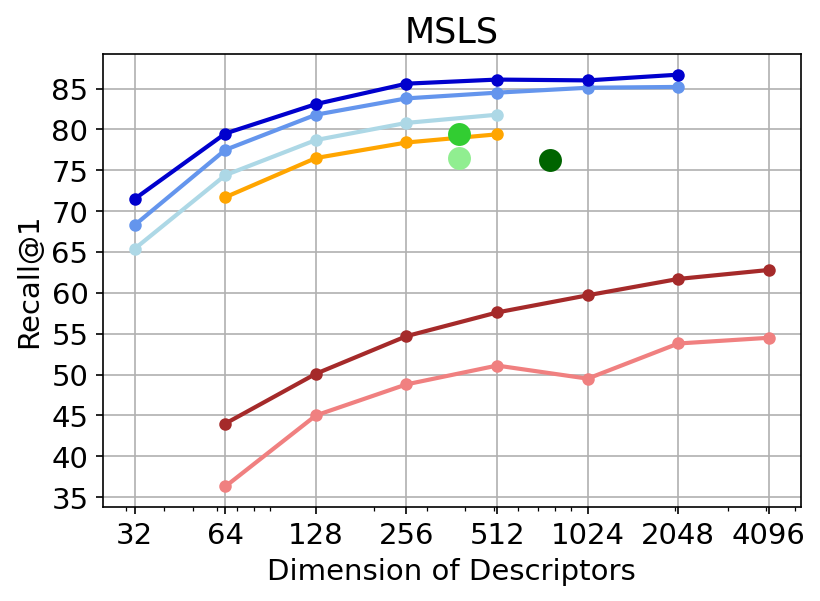}
        \end{subfigure}
    \end{minipage}
    \begin{minipage}{.32\textwidth}
        \begin{subfigure}{\textwidth}
            \includegraphics[width=\textwidth]{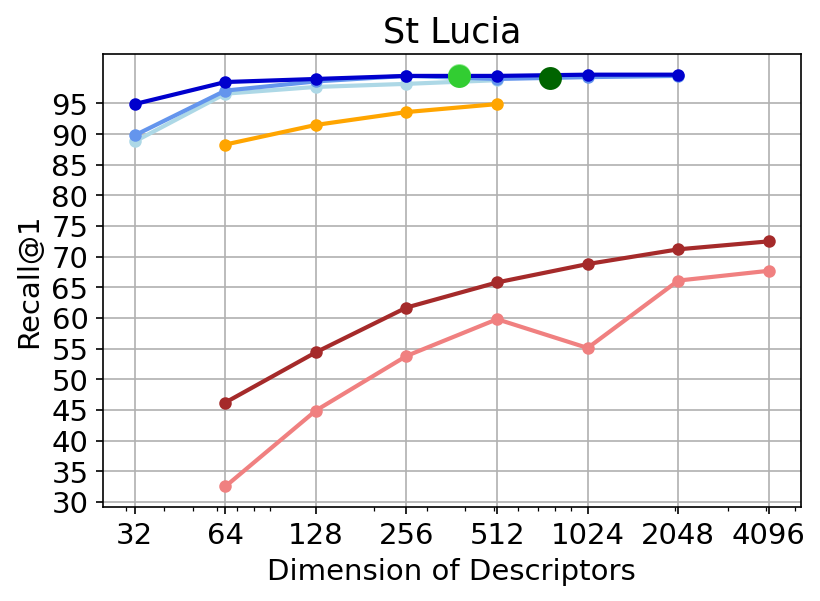}
        \end{subfigure}
        \centering
        \begin{subfigure}{\textwidth}
            \vspace{3mm}
            \centering
            \includegraphics[width=3.3cm, height=3.2cm]{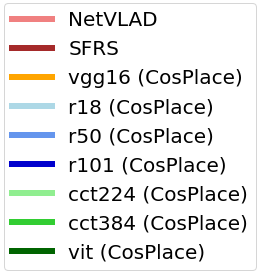}
            \vspace{2mm}
        \end{subfigure}
    \end{minipage}
    \caption{
    \textbf{Further results on backbones and descriptors dimensionality.}
    Results on a number of datasets of {\our} using different backbones and dimensionalities, compared with SFRS and NetVLAD trained on Pitts30k.
    }
    \label{fig:backbones}
\end{figure*}

Results from \cref{fig:backbones} clearly show that {\our} presents encouraging results regardless of the depth of the backbone, and we argue that future works should focus on more modern architectures, such as the ResNets, which are generally faster, lighter and achieve comparable or better results than the commonly used VGG-16.
We also see that {\our} is able to reach remarkable recalls and robustness with very low dimensions; for example, we see that any 128-D architecture trained with {\our} outperforms 4096-D NetVLAD (which is trained on Pitts30k) on any test dataset.

While transformers achieve lower results, we want to point out that we used the same hyperparameters for all experiments (\eg, same learning rate and optimizer), and we believe that performing a proper hyperparameter tuning independently for each backbone can increase the results shown in \cref{fig:backbones}, at the cost of a large number of experiments.
Moreover, while we used a resolution of $512 \times 512$ for CNNs, transformers require a smaller size, respectively $224 \times 224$ for ViT and CCT224, and $384 \times 384$ for CCT384, and this can provide a further explanation of the lower results with transformers.

\begin{table*}
\begin{adjustbox}{width=\textwidth}
\centering
\begin{tabular}{lcccccccccccccccccccccc}
\toprule
\multicolumn{1}{l}{\multirow{2}{*}{Method}} & \multicolumn{1}{c}{\multirow{2}{*}{Desc. dim.}} & \multicolumn{1}{c}{\multirow{2}{*}{Train set}} & \multicolumn{2}{c}{Pitts250k} & & \multicolumn{2}{c}{Pitts30k} & & \multicolumn{2}{c}{Tokyo 24/7} & & \multicolumn{2}{c}{MSLS} & & \multicolumn{2}{c}{St Lucia} \\
\cline{4-5} \cline{7-8} \cline{10-11} \cline{13-14} \cline{16-17}
\multicolumn{3}{c}{}
& R@1  & R@5  & & R@1  & R@5  & & R@1  & R@5 &
& R@1  & R@5  & & R@1  & R@5 \\
\hline
GeM \cite{Radenovic_2019_gem} &   512 & Pitts30k
& 75.3\footnotesize{$\pm$ 0.2} & 88.4\footnotesize{$\pm$ 0.3} &
& 77.9\footnotesize{$\pm$ 0.4} & 90.5\footnotesize{$\pm$ 0.3} &
& 46.4\footnotesize{$\pm$ 0.9} & 65.3\footnotesize{$\pm$ 0.7} &
& 51.8\footnotesize{$\pm$ 0.9} & 64.4\footnotesize{$\pm$ 0.9} &
& 59.9\footnotesize{$\pm$ 1.6} & 76.3\footnotesize{$\pm$ 2.0} \\
GeM \cite{Radenovic_2019_gem} &   512 & MSLS
& 65.3\footnotesize{$\pm$ 1.2} & 81.0\footnotesize{$\pm$ 1.6} &
& 71.6\footnotesize{$\pm$ 2.1} & 85.1\footnotesize{$\pm$ 1.9} &
& 44.9\footnotesize{$\pm$ 1.7} & 62.6\footnotesize{$\pm$ 1.2} &
& 66.7\footnotesize{$\pm$ 0.7} & 78.9\footnotesize{$\pm$ 0.5} &
& 84.6\footnotesize{$\pm$ 1.1} & 93.3\footnotesize{$\pm$ 0.7} \\
GeM \cite{Radenovic_2019_gem} &   512 & {\ourD}*
& 64.7\footnotesize{$\pm$ 0.8} & 81.4\footnotesize{$\pm$ 0.8} &
& 67.8\footnotesize{$\pm$ 0.6} & 83.6\footnotesize{$\pm$ 0.7} &
& 37.9\footnotesize{$\pm$ 2.3} & 51.0\footnotesize{$\pm$ 2.1} &
& 46.8\footnotesize{$\pm$ 2.1} & 58.1\footnotesize{$\pm$ 1.2} &
& 68.5\footnotesize{$\pm$ 2.4} & 82.7\footnotesize{$\pm$ 1.8} \\
NetVLAD \cite{Arandjelovic_2018_netvlad} &   512 & Pitts30k
& 83.7\footnotesize{$\pm$ 0.3} & 92.8\footnotesize{$\pm$ 0.1} &
& 83.0\footnotesize{$\pm$ 0.2} & 92.6\footnotesize{$\pm$ 0.3} &
& 52.6\footnotesize{$\pm$ 1.1} & 70.9\footnotesize{$\pm$ 1.2} &
& 51.1\footnotesize{$\pm$ 1.0} & 63.5\footnotesize{$\pm$ 0.8} &
& 59.8\footnotesize{$\pm$ 0.5} & 74.5\footnotesize{$\pm$ 1.2} \\
NetVLAD \cite{Arandjelovic_2018_netvlad} &   512 & MSLS
& 74.6\footnotesize{$\pm$ 1.3} & 86.8\footnotesize{$\pm$ 1.2} &
& 77.0\footnotesize{$\pm$ 0.9} & 88.6\footnotesize{$\pm$ 1.2} &
& 50.5\footnotesize{$\pm$ 2.2} & 65.1\footnotesize{$\pm$ 1.7} &
& 72.6\footnotesize{$\pm$ 0.5} & 83.0\footnotesize{$\pm$ 0.3} &
& 92.6\footnotesize{$\pm$ 0.6} & 97.1\footnotesize{$\pm$ 0.4} \\
NetVLAD \cite{Arandjelovic_2018_netvlad} &   512 & {\ourD}*
& 77.5\footnotesize{$\pm$ 0.5} & 88.5\footnotesize{$\pm$ 0.2} &
& 79.7\footnotesize{$\pm$ 0.3} & 90.0\footnotesize{$\pm$ 0.4} &
& 53.0\footnotesize{$\pm$ 0.9} & 70.2\footnotesize{$\pm$ 0.5} &
& 53.1\footnotesize{$\pm$ 3.2} & 64.2\footnotesize{$\pm$ 2.2} &
& 78.7\footnotesize{$\pm$ 1.6} & 88.1\footnotesize{$\pm$ 1.8} \\
CRN \cite{Kim_2017_crn} &   512 & Pitts30k
& 84.6\footnotesize{$\pm$ 0.6} & 93.6\footnotesize{$\pm$ 0.3} &
& 84.3\footnotesize{$\pm$ 0.2} & 92.7\footnotesize{$\pm$ 0.2} &
& 53.4\footnotesize{$\pm$ 0.5} & 70.6\footnotesize{$\pm$ 0.8} &
& 54.1\footnotesize{$\pm$ 0.6} & 66.1\footnotesize{$\pm$ 0.6} &
& 56.6\footnotesize{$\pm$ 2.7} & 75.5\footnotesize{$\pm$ 2.9} \\
APANet \cite{Zhu_2018_apanet} \textdagger & 512 & Pitts30k
& 83.7 & 92.6 & &  -   &  -   & & 67.0 & 81.0 &
&  -   &  -   & &  -   &  -   \\
SARE \cite{Liu_2019_sare} &   512 & Pitts30k
& 84.3\footnotesize{$\pm$ 0.7} & 92.6\footnotesize{$\pm$ 0.4} &
& 84.7\footnotesize{$\pm$ 0.7} & 92.6\footnotesize{$\pm$ 0.5} &
& 62.0\footnotesize{$\pm$ 0.6} & 74.9\footnotesize{$\pm$ 0.4} &
& 55.8\footnotesize{$\pm$ 3.3} & 67.8\footnotesize{$\pm$ 3.3} &
& 63.4\footnotesize{$\pm$ 3.0} & 79.0\footnotesize{$\pm$ 1.9} \\
SFRS \cite{Ge_2020_sfrs} &   512 & Pitts30k
& 87.1\footnotesize{$\pm$ 0.4} & 94.6\footnotesize{$\pm$ 0.2} &
& 86.4\footnotesize{$\pm$ 0.5} & 93.8\footnotesize{$\pm$ 0.2} &
& 66.7\footnotesize{$\pm$ 1.0} & 79.6\footnotesize{$\pm$ 0.9} &
& 57.6\footnotesize{$\pm$ 1.1} & 68.9\footnotesize{$\pm$ 1.0} &
& 65.8\footnotesize{$\pm$ 3.1} & 80.1\footnotesize{$\pm$ 2.3} \\
SRALNet \cite{Peng_2021_sralNet} \textdagger & 512 & Pitts30k
& 84.8 & 93.5 & &  -   &  -   & & 60.6 & 76.5 &
&  -   &  -   & &  -   &  -   \\
APPSVR \cite{Peng_2021_appsvr} \textdagger & 512 & Pitts30k
& 85.3 & 94.0 & &  -   &  -   & & 62.0 & 76.5 &
&  -   &  -   & &  -   &  -   \\
\hline
\textbf{{\our} (Ours)} &   512 & {\ourD}
& 89.3\footnotesize{$\pm$ 0.2} & \textbf{96.2\footnotesize{$\pm$ 0.3}} &
& 88.5\footnotesize{$\pm$ 0.1} & \textbf{94.5\footnotesize{$\pm$ 0.2}} &
& \textbf{82.2\footnotesize{$\pm$ 0.5}} & \textbf{88.9\footnotesize{$\pm$ 0.9}} &
& \textbf{79.6\footnotesize{$\pm$ 0.5}} & \textbf{87.2\footnotesize{$\pm$ 0.4}} &
& \textbf{94.1\footnotesize{$\pm$ 0.8}} & 97.4\footnotesize{$\pm$ 0.1} \\
\bottomrule
\end{tabular}
\end{adjustbox}
\caption{\textbf{Comparisons of various methods on popular datasets with 512-D descriptors.} This table is the equivalent of Tab. 3 in the main paper, but here all descriptors have the same dimensionality.
}
\label{tab:comp_all_ds_low_dim}
\vspace{-0.3cm}
\end{table*}

\begin{table}
\begin{adjustbox}{width=\linewidth}
\centering
\begin{tabular}{lccccccccc}
\toprule
\multicolumn{1}{l}{\multirow{2}{*}{Method}} & \multicolumn{1}{c}{\multirow{2}{*}{Desc. dim.}} & \multicolumn{1}{c}{\multirow{2}{*}{Train set}} & \multicolumn{3}{c}{{\ourD} test v1} & & \multicolumn{3}{c}{{\ourD} test v2} \\
\cline{4-6} \cline{8-10}
\multicolumn{3}{c}{}
& R@1  & R@5  & R@10  & & R@1  & R@5  & R@10 \\
\hline
GeM           &   512 & Pitts30k
& 21.7 & 30.3 & 34.4 & & 43.1 & 63.7 & 69.2 \\
GeM           &   512 & MSLS
&  8.1 & 15.6 & 20.2 & & 29.3 & 46.3 & 53.8 \\
GeM           &   512 & {\ourD}*
&  9.8 & 17.6 & 21.2 & & 34.8 & 55.5 & 63.0 \\
NetVLAD       &   512 & Pitts30k
& 27.4 & 38.1 & 43.6 & & 66.7 & 79.3 & 82.9 \\
NetVLAD       &   512 & MSLS
& 14.5 & 21.0 & 28.9 & & 40.5 & 59.7 & 64.4 \\
NetVLAD       &   512 & {\ourD}*
& 25.4 & 32.9 & 40.5 & & 66.9 & 78.6 & 82.8 \\
CRN           &   512 & Pitts30k
& 31.4 & 43.0 & 49.7 & & 68.2 & 81.3 & 83.3 \\
SARE          &   512 & Pitts30k
& 30.8 & 42.1 & 46.5 & & 69.2 & 81.1 & 83.1 \\
SFRS          &   512 & Pitts30k
& 35.6 & 49.7 & 54.8 & & 78.1 & 88.5 & 91.3 \\
\hline
\textbf{{\our} (Ours)} & 512 & {\ourD}
& \textbf{65.1} & \textbf{73.6} & \textbf{77.6} & & \textbf{83.4} & \textbf{92.1} & \textbf{94.8} \\
\bottomrule
\end{tabular}
\end{adjustbox}
\caption{\textbf{Comparisons of various methods on {\ourD} test v1 and {\ourD} test v2 with 512-D descriptors.} This table is the equivalent of Tab. 4 in the main paper.
}
\label{tab:comp_sf_xl_low_dim}
\vspace{-0.3cm}
\end{table}

\subsubsection{Comparison with other methods using same descriptors dimensionality.}
\label{sec:comparisons_with_pca}
Given that {\our} uses much lower dimensionality of descriptors, in \cref{tab:comp_all_ds_low_dim} and \cref{tab:comp_sf_xl_low_dim} we report the equivalent experiments of Tab. 3 and Tab. 4 of the main paper, but using the same (512) dimensionality for all methods.
We can see that in this scenario, the advantages of {\our} w.r.t. previous works are even more noticeable.


\subsubsection{Comparison with models trained on Google Landmark.}
In \cref{tab:cosPlace_vs_gldv1}, we compare models trained using {\our} on SF-XL with models trained on two popular landmark retrieval (LR) datasets, namely the Google Landmark (GLD) and SfM120k \cite{Radenovic_2019_gem}.
Models trained on GLD and SfM120k are downloaded from the official repository of \cite{Radenovic_2019_gem} \footnote{{\small{\url{https://github.com/filipradenovic/cnnimageretrieval-pytorch}}}}, which relied on a triplet loss for training.
{\our} can't be used on such landmark retrieval datasets, as they lack GPS coordinates and heading labels.

Note that these experiments are not aimed at providing a rigorous comparison of {\our} vs triplet losses or SF-XL vs standard retrieval datasets, given that the underlying tasks (\ie VG and LR) present many differences; we just want to provide an intuition on how popular models trained for LR fare on VG datasets.

\begin{table}[!ht]
  \centering
  \resizebox{\columnwidth}{!}{
  \begin{tabular}{lllllllllllll}
    \toprule
    Training Dataset & Backbone & Pitts250k & Pitts30k & Tokyo 24/7 & MSLS & St Lucia \\
    \midrule
    SfM120k &ResNet-50  & 84.5 & 83.4 & 75.2 & 64.5 & 73.9 \\
    GLD   &ResNet-50  & 85.8 & 84.1 & 77.8 & 69.5 & 77.3 \\
    SF-XL   &ResNet-50  & \textbf{92.3} & \textbf{90.9} & \textbf{87.3} & \textbf{85.2} & \textbf{99.5} \\
    \midrule
    SfM120k &ResNet-101 & 85.0 & 83.9 & 77.5 & 64.7 & 76.3 \\
    GLD   &ResNet-101 & 86.9 & 85.1 & 77.8 & 72.4 & 83.4 \\
    SF-XL   &ResNet-101 & \textbf{91.8} & \textbf{90.5} & \textbf{88.9} & \textbf{86.7} & \textbf{99.7} \\
    \bottomrule
  \end{tabular}}
   \caption{\textbf{Comparison with models trained on large landmark retrieval datasets.} 
   The models trained on {\ourD} is trained with {\our}, while models trained on GLD and SfM120k rely on a triplet loss.
   All models are equivalent (\ie ResNets followed by a GeM pooling and a fully connected layer with output dimensionality 512).
   }
  \label{tab:cosPlace_vs_gldv1}
\end{table}


\begin{figure*}
    \centering
    \includegraphics[width=0.99\textwidth]{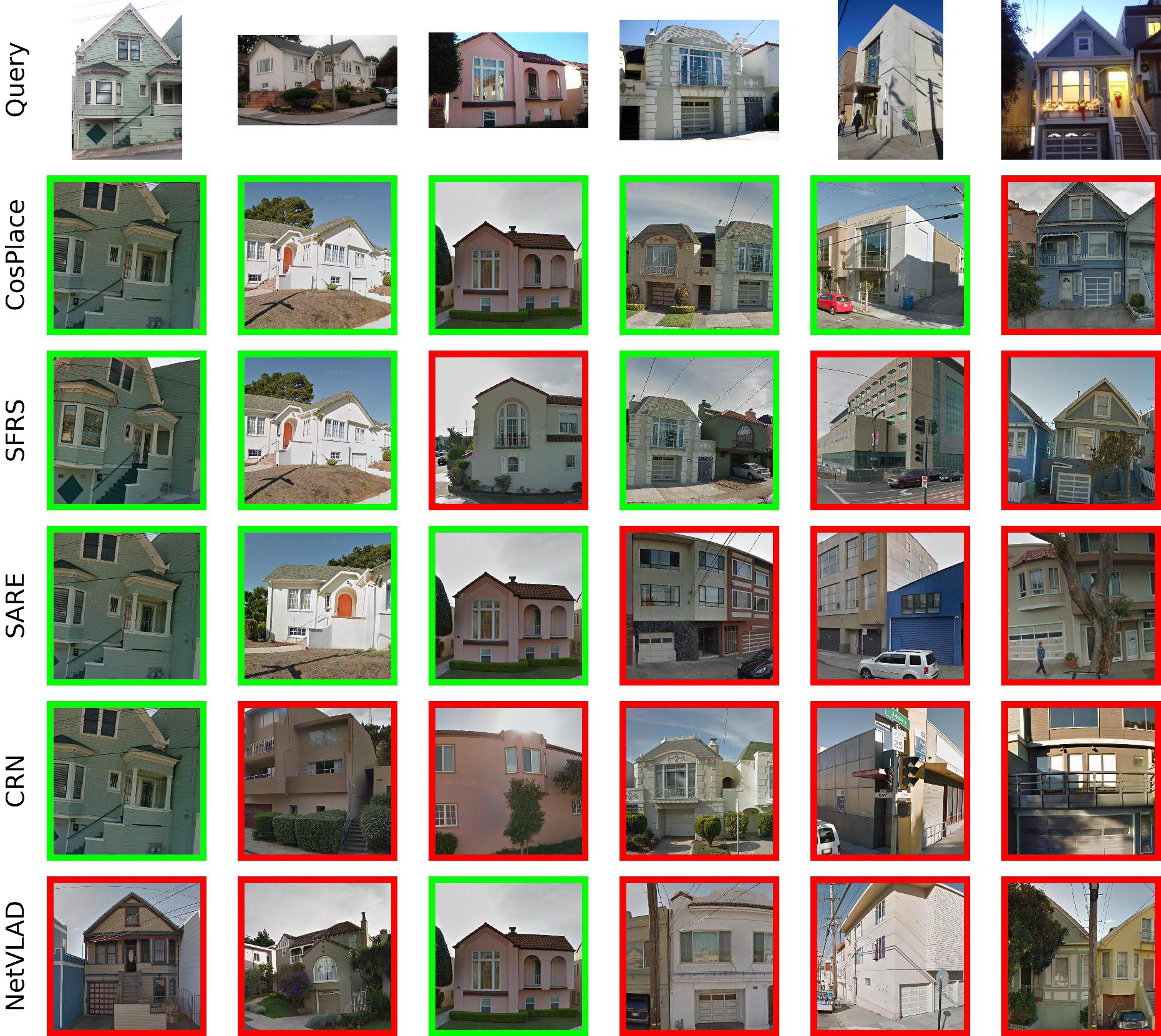}
    \caption{
    \textbf{Qualitative comparisons of retrieved images for a number of methods.}
    }
    \label{fig:grid}
\end{figure*}

\subsubsection{Comparison with other methods: qualitative results.}
\Cref{fig:grid} shows some qualitative results of retrieved images with {\our} compared to previous SOTA methods such as NetVLAD \cite{Arandjelovic_2018_netvlad}, CRN \cite{Kim_2017_crn}, SARE \cite{Liu_2019_sare} and SFRS \cite{Ge_2020_sfrs}.

\end{document}